\documentclass{article}
\usepackage{graphicx} 
\usepackage{amsfonts,amsmath}
\usepackage{graphicx}
\usepackage{hyperref}
\usepackage{algorithm}
\usepackage{algorithmic}
\usepackage{mathtools}

\usepackage{caption}
\usepackage{newfloat}

\newtheorem{theorem}{Theorem}
\newtheorem{example}{Example}
\newtheorem{practice}{Practice}

\usepackage{amsopn}

\usepackage{hyperref}
\usepackage{color}
\usepackage{listings}
\usepackage[caption=false]{subfig}

\usepackage[margin=0.7in]{geometry} 
\usepackage[many]{tcolorbox}    	

\definecolor{main}{HTML}{5989cf}    
\definecolor{sub}{HTML}{cde4ff}     

\tcbset{%
    sharp corners,
    colback = white,
    before skip = 0.2cm,    
    after skip = 0.5cm      
}                           

\newtcolorbox{boxK}{%
    sharpish corners, 
    boxrule = 0pt,
    toprule = 4.5pt, 
    enhanced,
    fuzzy shadow = {0pt}{-2pt}{-0.5pt}{0.5pt}{black!35} 
}

\DeclareFloatingEnvironment{boxes} 
\newcommand{\boxref}[1]{\hyperref[{#1}]{Box~\ref*{#1}}}

\DeclareFixedFont{\ttb}{T1}{txtt}{bx}{n}{10} 
\DeclareFixedFont{\ttm}{T1}{txtt}{m}{n}{10}  

\definecolor{deepblue}{rgb}{0,0,0.5}
\definecolor{deepred}{rgb}{0.6,0,0}
\definecolor{deepgreen}{rgb}{0,0.5,0}

\lstdefinestyle{custompython}{%
    language=Python,
    numbers=left,                    
    numbersep=5pt,                   
    captionpos=b,
    commentstyle=\color{gray}, 
    numberstyle=\tiny\color{gray}, 
    basicstyle=\footnotesize,
    morekeywords={self},              
    keywordstyle=\ttb\color{deepblue},
    emph={MyClass,__init__},          
    emphstyle=\ttb\color{deepred},    
    stringstyle=\color{deepgreen},
    showstringspaces=false
}

\newcommand{\X}{{\bf x}}
\newcommand{\Y}{{\bf y}}
\newcommand{\Ti}{{\bf t}}
\newcommand{\FHL}{FC\textsuperscript{h}}
\newcommand{\FOL}{FC\textsuperscript{out}}
\newcommand{\FINL}{FC\textsuperscript{input}}
\newcommand{\BV}{IBVP\@ }

\DeclareMathOperator*{\argmin}{arg\,min}

\makeatletter
\renewcommand{\maketitle}{\bgroup\setlength{\parindent}{0pt}
\begin{flushleft}
  \textbf{\@title}

  \@author
\end{flushleft}\egroup
}
\makeatother

\title{{\Large Practical Aspects on Solving Differential Equations Using Deep Learning: A Primer}}

\author{Georgios Is. Detorakis \\ International Centre for Neuromorphic Systems \\ Department of Computer Science \\ University of Manchester \\ 
  \url{georgios.detorakis@manchester.ac.uk}}
\date{}

\begin{document}

\maketitle

\begin{abstract}

Deep learning is now common across many scientific fields, including the study of partial differential equations. This article provides a brief, accessible introduction to core deep learning concepts, including neural networks, backpropagation, and the universal approximation theorem. It mainly covers how to use deep learning in solving differential equations. The article aims to help undergraduate and graduate students in mathematics, physics, and related areas learn how to use Deep Learning to solve partial differential equations. Instructors in mathematics or physics can also use this article to introduce students to Deep Galerkin method and scientific deep learning. We focus on key questions: What is deep learning, and how can it help solve mathematical or physical problems? How can you implement a neural network and choose the right numerical method to solve differential equations? How do you select the best hyperparameters? How can you improve accuracy and speed up convergence? We should mention that all the problems in this article can be solved on a machine without a GPU, so any student can follow the presented methodology. 
\end{abstract}

\paragraph{Keywords}
Deep Neural Networks, Deep Galerkin Method, Partial and Ordinary Differential Equations, Systems of Ordinary Differential Equations,
Pytorch, Scientific Machine Learning

\paragraph{MSCcodes}
97P80, 97M10, 65M99, 65P99, 65L99

\tableofcontents

\section{Motivation}

Deep learning (DL)~\cite{lecun:2015} has advanced significantly in recent years, providing algorithms that can solve complex problems ranging from image classification~\cite{byerly:2022} to playing games such as Go~\cite{schrittwieser:2020} at a human level, and even better. More recent advances include the development of Large Language Models (LLMs), which are foundation deep learning models (\emph{i.e.}, they usually have billions of parameters) such as ChatGPT~\cite{stiennon:2020,gao:2023} and Llama~\cite{touvron:2023}, which have been trained on enormous data sets. 

Moreover, deep learning has found many applications in various scientific disciplines, including mathematics and physics. Recent years have seen a steady increase in the scientific literature focused on deep learning in these areas. Scientific deep learning is used in various contexts, such as solving partial differential equations~\cite{lagaris:1998,sirignano:2018}, optimizing energy functionals in materials science~\cite{mema:2025}, and constructing neural operators~\cite{li:2021,li:2020}.

To this end, this article aims to introduce students of mathematics, physics, or related fields to the essential concepts of scientific deep learning. As a primer, it presents no new ideas but instead shows how deep neural networks can be used in fields such as mathematics and physics to solve problems primarily related to the numerical solution of partial differential equations. This article also addresses the tuning of hyperparameters, which can often be tedious and time-consuming. The main motivation for this concise article is to present the core ideas of deep learning, provide simple examples, and apply one of the most widely used and fundamental methods of scientific deep learning to a simple problem, from the basic formulation through implementation and tuning. 

The rest of this article is organized as follows: First, we introduce the basic terminology and notation used by the deep learning community. We introduce the basic data structures used in Pytorch, and we provide a brief introduction to feed-forward neural networks, DGM networks (\emph{i.e.}, a type of recurrent neural networks proposed in~\cite{sirignano:2018}), and residual neural networks (ResNets). We then introduce the universal approximation theorem and its implications for scientific deep learning. Moreover, we provide an intuitive example that clearly illustrates the theorem's main idea. In turn, we present the finite differences method and the Galerkin method, and then connect them to deep learning, showing how we can use deep neural networks to implement a Deep Galerkin method and solve problems in the fields of partial and ordinary differential equations,
such as the $1$-d heat equation. Finally, we demonstrate how we can apply the deep Galerkin method on ordinary differential equations (ODEs), systems of ODEs, and integral equations.

\begin{tcolorbox}[title={\bf Brief History of Scientific Machine Learning}]
    Early neural-network solvers [45, 69, 50] relied on Hopfield networks or combined neural networks with spline approximations. The work of Lagaris et al. [42] introduced the key idea of constructing a loss function from the differential operator. With this approach, a neural network approximates the solution by minimizing that loss—an idea that underpins most modern approaches. Other important works that pioneered the field of scientific deep learning include, but are not limited to: the deep splitting scheme for semilinear PDEs [4], Kolmogorov PDE solvers [5, 60], backward stochastic differential equation methods [28, 27], and physics-informed neural networks (PINNs) [57, 39]. The reader is referred to [6] for a comprehensive survey and to [10] for an accessible overview of three representative approaches. It is worth noting that deep-learning-based solvers have known limitations. When training data are sparse, PINNs and related methods may produce solutions that diverge from those of traditional solvers [14, 15]. The DG method is therefore best viewed as a complement to, rather than a replacement for, classical methods such as finite elements [43].
\end{tcolorbox}
\noindent\begin{minipage}{\textwidth}
\label{box:history}
\end{minipage}

\section{Deep Learning}
\label{section:neural_nets}

To begin, we introduce the concept of a neural network. Neural networks, or artificial neural
networks (ANNs), are at the heart of deep learning~\cite{goodfellow:2016}. In the literature, we find a wide variety of neural networks, like feed-forward networks, convolutional neural networks~\cite{goodfellow:2016}, recurrent neural networks~\cite{rumelhart:1986}, long short-term memory (LSTM) networks~\cite{hochreiter:1997}, and ResNets~\cite{he:2016}. Of course, there are many more, but they are out of the scope of this module (see~\cite{bishop:2006,goodfellow:2016,burkov:2019,geron:2022,higham:2019}). Now we will turn our focus to three types of neural networks: (i) feed-forward neural networks, (ii) DGM networks, a type of recurrent neural network, and (iii) residual neural networks (ResNets). 

\subsection{Feed-forward Neural Networks}
\label{section:mlp}
Let's delve into the practical application of feed-forward neural networks, also known as multi-layer perceptrons (MLPs). These networks are not just theoretical concepts but powerful tools that map an input ${\bf x} \in \mathbb{R}^m$ to an output ${\bf y} \in \mathbb{R}^p$. In simpler terms, a neural network defines a mapping ${\bf y} = \mathcal{F}({\bf x}; \pmb{\theta})$ from an input to an output, making it a key component in machine learning and artificial intelligence. For instance, an MLP with three hidden layers, one input layer, and one output layer can be defined by four compositions as:
\begin{align}
	{\bf y} &= \mathcal{F}({\bf x}; \pmb{\theta}) = f^{(4)} \circ f^{(3)} \circ f^{(2)} \circ f^{(1)}({\bf x}) = f^{(4)}(f^{(3)}(f^{(2)}(f^{(1)}({\bf x})))). \nonumber
\end{align}
Here, $f_1$ maps the input to the first hidden layer; $f_2$ maps the first hidden to the second; and so on. Expanding the composition with affine
transformations gives:
\begin{align}
	 {\bf y} &= \mathcal{F}({\bf x}; \pmb{\theta}) = {\bf W}_4\cdot g_3
	 \bigg({\bf W}_3 \cdot g_2\Big({\bf W}_2 \cdot g_1({\bf W}_1 \cdot {\bf x} + {\bf b}_1) + {\bf b}_2\Big) + {\bf b}_3\bigg) + {\bf b}_4. \nonumber
\end{align}
In this case, the output layer $L$ has no activation function. Moreover, we can write the activity of one layer in the form ${\bf y}_l = f_l({\bf z}) = g_l({\bf W}_l \cdot {\bf z} + {\bf b}_l)$, where ${\bf W}_l$ are the weights that connect neurons from layer $l-1$ with neurons in layer $l$ and ${\bf b}_l$ are the biases of the neurons at layer $l$, with $l = 1, \ldots, k$.
The function $g(x)$ is called the activation function and may be (i) a Sigmoid $g(x) = \frac{1}{1 + \exp(-x)}$, (ii) a hyperbolic tangent $g(x) = \tanh(x)$, (iii) a rectified linear unit (ReLU) $g(x) = \max\{x, 0\}$, any custom activation, or any function found in the PyTorch documentation page\footnote{\url{https://pytorch.org/docs/stable/nn.html\#\#non-linear-activations-weighted-sum-nonlinearity}}.
Typically, we will group all the neural network parameters, weights, ${\bf W}_i$, and biases, ${\bf b}_i$, in one symbol, $\pmb{\theta}$, so when we want to refer to the parameters of a neural network, we will use that notation. Of course, the number of hidden layers and the number of neurons (units) within a layer can vary. We will call a neural network deep when it has more than two hidden layers~\cite{goodfellow:2016,burkov:2019}. Moreover, each layer in a feed-forward neural network is called a fully connected (FC) or linear layer, and this is the term PyTorch uses. 
From now on, we will refer to a fully connected layer as FC, and more precisely, input layers as \FINL, hidden layers as \FHL, and the output ones as \FOL\@.
\begin{figure}
    \centering
    \includegraphics[width=1\textwidth]{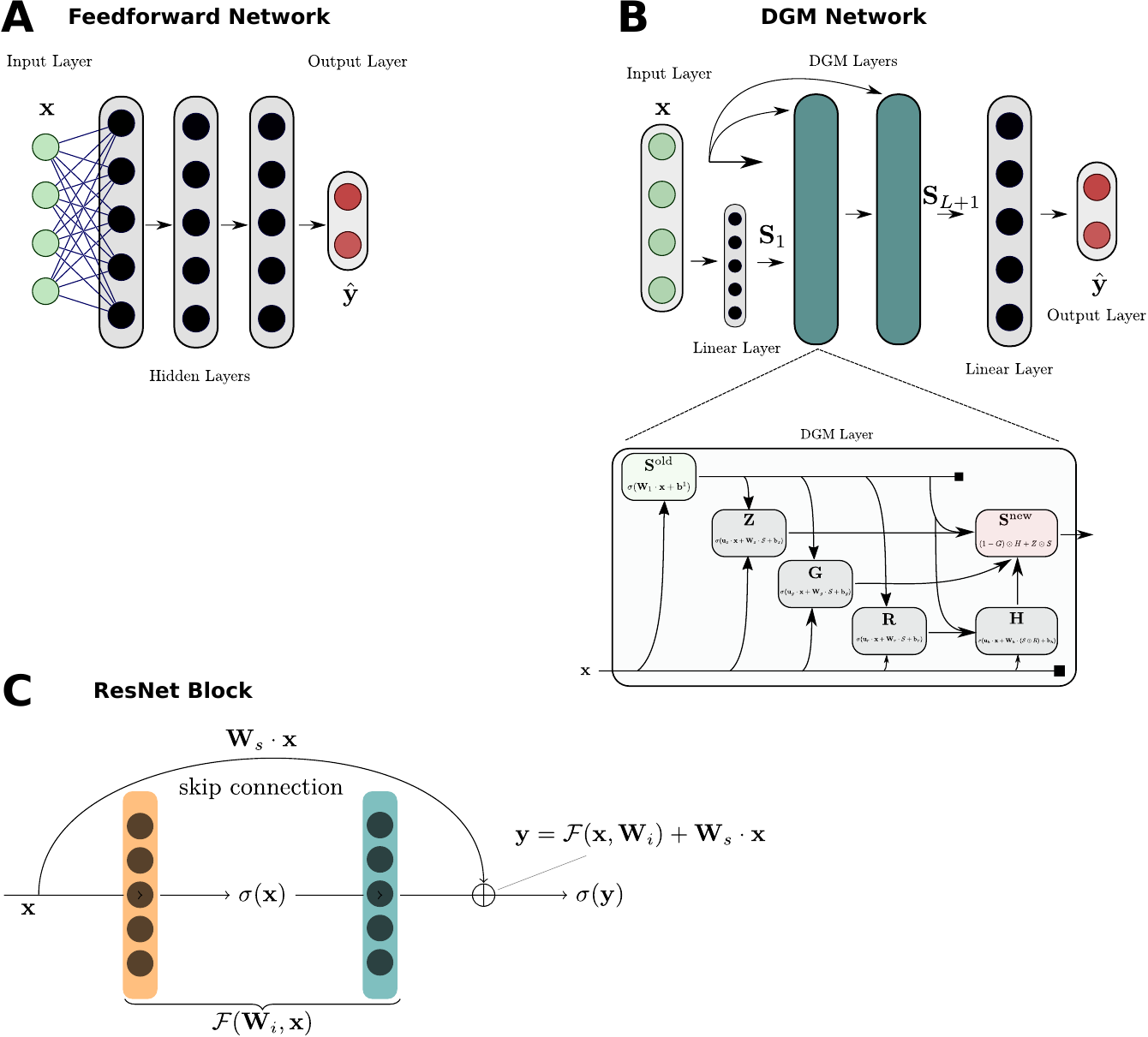}
    \caption{{\bfseries \sffamily Neural Network Architectures}.
    {\bfseries \sffamily A} A feed-forward neural network with three hidden layers (gray color),
    one input (green color), and one output (red color) layer. The connections from the input to the first
    hidden layer are visible in the graph. 
    {\bfseries \sffamily B} A DGM neural network with two DGM layers (teal color) and two fully
    connected (linear) layers (gray color). The inset shows the flow of information and the
    transformations within a DGM layer.
    {\bfseries \sffamily C} A residual block of a ResNet.
    The input is distributed to the residual mapping (orange and teal colors) and
    to the output of the block via a skip connection. }
    \label{fig:nn_architecture}
\end{figure}
In Figure~\ref{fig:nn_architecture} {A}, we can see a typical feed-forward neural network, where the input (green) and output (red) layers are visible, as well as the hidden layers shown as gray boxes. The connectivity matrix from the input to the first hidden layer is provided; the remaining connections are omitted for simplicity. 

\subsubsection{Pytorch Nomenclature}
\label{section:terminology}
\begin{figure}[!htpb]
    \centering
    \includegraphics[width=.7\textwidth]{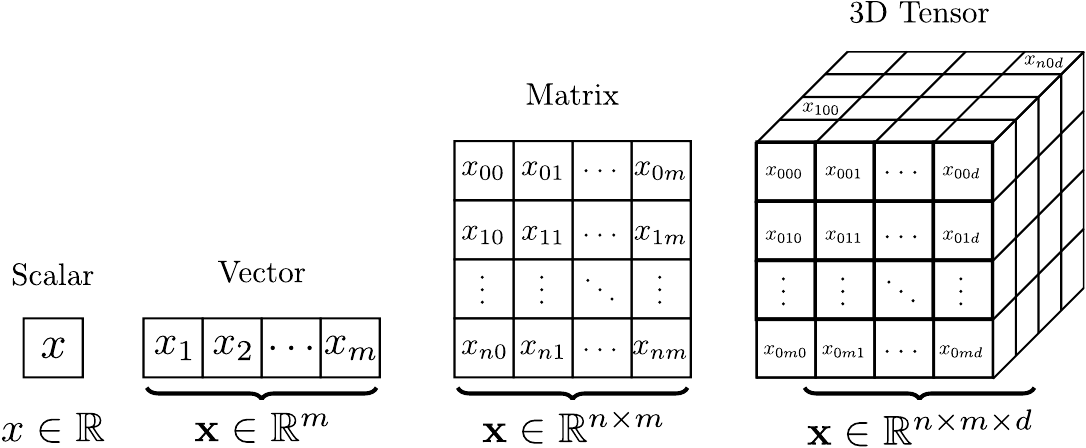}
    \caption{{\bfseries \sffamily Visual representation of Pytorch tensors}. From left to right
    we see a scalar, $x \in \mathbb{R}$, a one-dimensional tensor (or vector) of dimension $m$, ${\bf x} \in \mathbb{R}^m$,
    a two-dimensional tensor (or matrix) ${\bf X} \in \mathbb{R}^{n \times m}$, and finally a three-dimensional tensor
    ${\bf X} \in \mathbb{R}^{n\times m \times d}$. The index starts from zero, following Python's convention. } 
    \label{fig:tensors}
\end{figure}

Since we rely on PyTorch to implement our algorithms and neural networks, we briefly introduce some basic terminology. Therefore, when we refer to a Pytorch {\bf Tensor}, we mean a data structure that holds information (\emph{i.e.}, data) along with the appropriate methods to perform various operations on it, for instance, to compute the mean or the standard deviation. Students who are familiar with Numpy can see that Pytorch Tensors are multidimensional Numpy arrays. At this point, we should distinguish Pytorch tensors from mathematical tensors, since the former are essentially multidimensional arrays. In addition, a PyTorch tensor (simply a tensor from now on) has a shape where the first dimension is usually the batch size (i.e., how many data samples the tensor contains), and the remaining dimensions hold the data sizes. For example, ten $3$-by-$3$ matrices can be stored in a tensor with shape $(10, 3, 3)$. In Figure~\ref{fig:tensors}, we show a schematic representation of tensors of different shapes.

Another frequently used term is mini-batch (or just batch). A mini-batch is a small subset of the training or test dataset, used when a dataset is split into smaller chunks. We also shuffle these smaller chunks. Mini-batches are crucial in stochastic gradient descent, as we will see later. 

\subsubsection{Pytorch Implementation of an MLP}
Like all modern deep learning libraries, PyTorch provides tools for building feed-forward neural networks and defining affine transformations easily. The reader can find a complete list on the PyTorch documentation page\footnote{\url{https://pytorch.org/docs/stable/nn.html\#linear-layers}}.
Moreover, Listing~\ref{list:mlp} shows a basic implementation of a Pytorch Multi-Layer Perceptron neural network with linear layers and $\tanh$ activation functions. We must define the layers and activation functions used in the \texttt{\_\_init\_\_} constructor method. The argument \texttt{input\_dim} defines the number of input units (independent variables such as $t$ and ${\bf x}$), and on the other hand, the argument \texttt{output\_dim} determines the number of units in the output layer (\emph{e.g.}, $y$). Moreover, it receives the number of layers (\texttt{num\_layers}) and the number of units in each hidden layer (\texttt{hidden\_size}). In this implementation, we assume that all hidden layers will have the same number of units. Keep in mind that when we use a neural network with one hidden layer, the number of layers must be set to 1; for two hidden layers, it must be set to 2, and so on.  In line $11$ of code listing~\ref{list:mlp}, we define the input layer; in lines $14$--$15$, the hidden layers; and in line $18$, the output layer. The class \texttt{Linear} of Pytorch is an implementation of an affine transformation (${\bf y} = {\bf W}\cdot {\bf x} + {\bf b}$). Notice that in line $14$, we use a \texttt{ModuleList} to instantiate the hidden layers. We do so because the number of hidden layers is not known \emph{a priori}; instead, it is passed as an argument to the constructor when we instantiate the \texttt{MLP} class. The \texttt{forward} method takes the input tensor as an argument (it might accept more arguments, as we will see later) and applies all the transformations and activation functions we have already defined. Essentially, this method runs the forward pass of our neural network, which is the part of the code that determines the structure of our neural network. When we use the batch normalization (\texttt{batch\_norm = True}), we eliminate the bias term from the previous Layer by setting \texttt{bias=False} in the constructor of the \texttt{Layer} class in Pytorch.

\begin{lstlisting}[style=custompython,label=list:mlp,caption={Pytorch
Implementation of a Multi-Layer Perceptron class.}]
class MLP(nn.Module):
    def __init__(self,
                 input_dim=2,
                 output_dim=1,
                 hidden_size=50,
                 num_layers=1,
                 batch_norm=False):
        super(MLP, self).__init__()

        # Input layer
        self.fc_in = nn.Linear(input_dim, hidden_size)

        # Hidden layers
        self.layers = nn.ModuleList([nn.Linear(hidden_size, hidden_size)
                                     for _ in range(num_layers)])

        # Output layer
        self.fc_out = nn.Linear(hidden_size, output_dim)

        # Non-linear activation function
        self.act = nn.Tanh()

        # Batch normalization
        if batch_norm:
            self.bn = nn.BatchNorm1d(hidden_size)
        else:
            self.bn = nn.Identity()
       
    def forward(self, x):
        out = self.act(self.bn(self.fc_in(x)))
                
        for i, layer in enumerate(self.layers):
            out = self.act(self.bn(layer(out)))
        out = self.fc_out(out)
        return out
\end{lstlisting}

\subsection{DGM neural network}
\label{section:dgm_net}

The second type of neural network commonly used for solving differential equations is the neural network proposed by~\cite{sirignano:2018}. We will call this neural network \emph{DGM}. A DGM is similar in concept to the Long Short-Term Memory (LSTM) neural networks introduced by
Hochreiter and Schmidhuber in~\cite{hochreiter:1997}, and to the Highway Networks introduced in~\cite{srivastava:2015}. LSTM is a special kind of recurrent neural network equipped with input, output, and forget gates, as well as a memory cell. LSTMs learn to balance memory and forget to learn long sequences, avoiding the vanishing gradient problem (we will, later,  explain the vanishing gradient problem after introducing gradient descent and backpropagation). 

A DGM layer receives an input ${\bf x}$ and a potential output of the previous DGM layer, and then it propagates the information through four affine transformations. The following equations describe the operations that take place within a DGM layer.  
\begin{align}
\label{eq:dgm_net}
\begin{split}
    S_1 &= \sigma({\bf W}_1 \cdot {\bf x} + {\bf b}^1 ), \\
    Z_l &= \sigma({\bf U}_{z, l} \cdot {\bf x} + {\bf W}_{z,l}\cdot S_l + {\bf b}_{z,l}), \\
    G_l &= \sigma({\bf U}_{g, l} \cdot {\bf x} + {\bf W}_{g,l}\cdot S_l + {\bf b}_{g,l}), \\
    R_l &= \sigma({\bf U}_{r, l} \cdot {\bf x} + {\bf W}_{r,l}\cdot S_l + {\bf b}_{r,l}), \\
    H_l &= \sigma({\bf U}_{h, l} \cdot {\bf x} + {\bf W}_{h,l}\cdot(S_l \odot R_l) + {\bf b}_{h, l}), \\
    S_{l+1} &= (1 - G_l) \odot H_l + Z_l \odot S_l, \\
    f(t;{\bf x}; \pmb{\theta}) &= {\bf W} \cdot S_{L+1} + {\bf b},
\end{split}
\end{align}
where $l=1,\ldots,L$ is the number of layers. Here, $\sigma$ is an activation function and $\odot$ is the element-wise multiplication (Hadamard product). In addition,${\bf U}$ represents the input to hidden weights, ${\bf W}$s are the hidden-to-hidden weithgs, and ${\bf b}$s are the biases.  
Note that each DMG layers receives two arguments, one is the input ${\bf x}$ and the other is the final state of the previous DGM layer. Now, we ask ourselves, ``What is the previous state of the first DMG layer?'' For the first DMG layer, we provide the input ${\bf x}$
and a projection of the input ${\bf x}$ to an affine transformation or a linear layer. Therefore, the input to the first DMG layer is: $({\bf x}, {\bf W}_{\text{in}}{\bf x} + {\bf b}_{\text{in}})$. 
Figure~\ref{fig:nn_architecture} {B} shows a DGM neural network with two DGM layers (teal boxes) and two fully connected (linear) layers (gray boxes). There is one input and one output layer. The inset illustrates the operations and information flow within a DGM layer. 

\subsubsection{Pytorch Implementation of a DGM}

Building upon this architecture, in Listing~\ref{list:dgm} we show the Pytorch implementation of a DGM layer proposed by~\cite{sirignano:2018}. The layer
notation follows Equation~\eqref{eq:dgm_net}; thus, the implementation is a straightforward 
translation of Equation~\eqref{eq:dgm_net} into PyTorch code. The \texttt{\_\_init\_\_} 
method receives the number of input features, \texttt{input\_dim}, and the number of 
hidden units, \texttt{hidden\_size}. 
Furthermore, notice that the \texttt{forward} method receives two arguments: the input \texttt{x} and the state of the previous DGM layer \texttt{s}.

    \begin{lstlisting}[style=custompython,label=list:dgm,caption={Pytorch Implementation of a DG Layer.}]
 class DGMLayer(nn.Module):
    def __init__(self, input_dim=1, hidden_size=50):
        super().__init__()

        self.Z_wg = nn.Linear(hidden_size, hidden_size)
        self.Z_ug = nn.Linear(input_dim, hidden_size, bias=False)

        self.G_wz = nn.Linear(hidden_size, hidden_size)
        self.G_uz = nn.Linear(input_dim, hidden_size, bias=False)

        self.R_wr = nn.Linear(hidden_size, hidden_size)
        self.R_ur = nn.Linear(input_dim, hidden_size, bias=False)

        self.H_wh = nn.Linear(hidden_size, hidden_size)
        self.H_uh = nn.Linear(input_dim, hidden_size, bias=False)

        # Non-linear activation function
        self.sigma = nn.Tanh()

    def forward(self, x, s):
        Z = self.sigma(self.Z_wg(s) + self.Z_ug(x))
        G = self.sigma(self.G_wz(s) + self.G_uz(x))
        R = self.sigma(self.R_wr(s) + self.R_ur(x))
        H = self.sigma(self.H_wh(s * R) + self.H_uh(x))
        out = (1 - G) * H + Z * s
        return out

        \end{lstlisting}

\subsection{Feed-forward ResNet}
\label{section:resnet}

The third and final neural network we introduce in this module is a residual neural network, or ResNet. ResNets were introduced by He et al., for image recognition~\cite{he:2016} in 2016. The main idea behind ResNets is residual learning, and at its heart, there is a residual block, which consists of either linear or convolutional layers and activation functions. A minimal block performs the following transformation:
\begin{align}
 {\bf y} = \Phi({\bf x}, \pmb{\theta}_{\text{res}}) + {\bf W}_s \cdot {\bf x}. \nonumber
\end{align}
Here, ${\bf x}$ and ${\bf y}$ are the block's input and output, respectively. The trainable mapping $\Phi({\bf x}, \pmb{\theta}_{\text{res}})$, called a residual, and corresponds to multiple convolutional layers in the original ResNet. Each residual block has its own learnable parameters $\pmb{\theta}_{\text{res}}$. The crucial feature is the skip connection, which propagates the input directly to the output without applying the residual mapping. If the input ${\bf x}$ and residual mapping share the same dimension, then ${\bf W}_s$ is the identity; otherwise, it must be learned so that the dimension of ${\bf W}_s \cdot {\bf x}$ matches that of the residual mapping. Figure~\ref{fig:nn_architecture}C shows a feed-forward ResNet block; for our purposes, we replace the convolutional layers with linear (fully connected) layers.

\subsubsection{Pytorch Implementation of a ResNet}

Now, we describe how to implement a simple ReNet block in PyTorch (the source code for a fully functional linear ResNet is available on the GitHub page for this module; see Motivation). Listing~\ref{list:resnet} shows the Pytorch code of a ResNet block. 

Let’s dive deeper into the implementation by first explaining the method \texttt{\_\_init\_\_}, which takes the number of units in the input and output layers as arguments, respectively. The argument \texttt{downsample} defines a transformation (\emph{i.e.}, ${\bf W}_s$) at skip connections, applied when the input's ${\bf x}$ dimensions mismatch the output of the residual block (see Figure~\ref{fig:nn_architecture} {C}). The argument \texttt{num\_features} indicates if the input tensor contains an extra dimension besides batch size and independent variables. The linear layers are defined using the Pytorch \texttt{Sequential} container in lines $10$--$14$. Each container instantiates a Pytorch linear layer followed by batch normalization (see Section~\ref{section:bn}). In the forward method
(lines $23$--$33$), we apply two linear layers (see lines $20$ and $21$ of code listing~\ref{list:resnet}) and linear rectifications. We check if \texttt{downsample} is not \texttt{None} (lines $23$ and $24$). If so, we apply the downsample transformation to ${\bf x}$; otherwise, our residual is the original input ${\bf x}$. Once we have the residual values, we add them to the output, apply linear rectification, and return the output.

\begin{lstlisting}[style=custompython,label=list:resnet,caption={Pytorch Implementation of a Linear ResNet Block.}]
class ResidualBlock(nn.Module):
    def __init__(self,
                 input_dim=2,
                 output_dim=1,
                 downsample=None,
                 num_features=None):
        super().__init__()
        self.downsample = downsample

        if num_features is None:
            self.num_features = output_dim
        else:
            self.num_features = num_features

        self.fc1 = nn.Sequential(nn.Linear(input_dim, output_dim, bias=False),
                                 nn.BatchNorm1d(self.num_features))

        self.fc2 = nn.Sequential(nn.Linear(output_dim, output_dim, bias=False),
                                 nn.BatchNorm1d(self.num_features))

        self.relu = nn.ReLU()

    def forward(self, x):
        residual = x
        out = self.relu(self.fc1(x))
        out = self.relu(self.fc2(out))

        if self.downsample is not None:
            residual = self.downsample(x)
        out = out + residual
        out = self.relu(out)

        return out
\end{lstlisting}

\subsection{Neural Network Initialization}
\label{section:init}

We have introduced the three major types of neural networks we need in this module. Now, we show how to initialize a neural network. Initialization plays a crucial role in neural network training, as it affects the optimization process, as we will see later. 

One might think that a simple way to initialize those parameters is to use a distribution, such as a uniform or a Gaussian, and draw the initial values from it. However, things, in reality, are not that simple. For instance, if we do not account for the mean and variance of a Gaussian distribution and we initialize our parameters with large or small values, we will most certainly affect the optimization process in a detrimental way.

To this end, we would like to
initialize our parameters so that the activation mean is zero and the variance remains the same across all layers. To do so, we can use methods such as Xavier’s initialization~\cite{glorot:2010}. In this case,
the weights and biases receive values based on:
\begin{align}
    \label{eq:xavier}
    \begin{split}
        {\bf W}_l &= \mathcal{N}(\mu = 0,
        \sigma^2= \frac{1}{k}), \\ {\bf b}_l &= {\bf 0},
    \end{split}
\end{align}
where ${\bf W}_l$ and ${\bf b}_l$ are the weights and the biases of layer $l$,
$\mathcal{N}$ is a normal distribution centered at zero (we can replace the normal 
distribution with a uniform one), and $k$ is the number
of inputs and outputs in the $l$ layer. Xavier initialization sets the initial weights across all layers to the same range. Moreover, it will accelerate the optimization and minimize potential oscillations around the minima.

It is worth mentioning that the authors in~\cite{he:2015} argue that the Xavier
initialization works well with activation functions such as a $\tanh$ or a sigmoid, 
but not with ReLU or LeakyReLU. Instead, they propose a slightly different initialization 
method, called He or Kaiming initialization, where they replace the variance in 
Equation~\eqref{eq:xavier} with $\frac{2}{k_{l-1}}$~\cite{kumar:2017,goodfellow:2016}. 
Pytorch supports Xavier initialization with normal and uniform distributions\footnote{\url{https://docs.pytorch.org/docs/2.12/nn.init.html}}.

\section{Backpropagation, Optimization \& Regularization}

\subsection{Optimization \& Gradient Descent}
\label{section:sgd}
So far we have seen three basic neural network types, how to properly initialize a neural
network, and how to implement them on Pytorch. 
However, we have not explained how to train a neural network. 
Therefore, now, we proceed in briefly explaining the training (learning) process of a neural network starting with the Gradient Descent. 

Essentially, we would like to find a set of parameters, $\pmb{\theta}$, that solve the 
problem we intend to use the neural network on. In other words, we are looking for the set 
of parameters $\pmb{\theta}$ that will minimize a loss function (or error function), 
$\mathcal{L}(\pmb{\theta};\mathcal{F}) $. The loss function takes the network's parameters as its main argument and receives the neural network as a parameter. It then evaluates an error on the output of the neural network, which will be used to search for an ``optimal’’ set of parameters $\pmb{\theta}^{\ast}$. Thus, we want to solve the following problem (we will omit all the parameters in loss function from now on):
\begin{align}
    \label{eq:argmin}
    \pmb{\theta}^{\ast} &= \argmin_{\pmb{\theta}} \{ \mathcal{L}(\pmb{\theta}). \}
\end{align}
One way to solve the optimization problem above (Equation~\eqref{eq:argmin}) is to use the Gradient Descent (GD), which is an optimization algorithm~\cite{ruder:2016} that starts with an initial guess of the parameters $\pmb{\theta}_0$ and uses the gradient at that point to obtain the direction of the most significant increase in the loss function. Then, it moves in the opposite direction to minimize the loss based on the following equation:
\begin{align}
    \label{eq:sgd}
    \pmb{\theta}_{n+1} = {\pmb \theta}_n - \eta \nabla_{\pmb{\theta}}
                    \mathcal{L}(\pmb{\theta}_n),
\end{align}
where $\eta$ is the learning rate. At each GD iteration we use the entire training data set at each iteration, as one ``mega''-batch. Therefore, the computational cost is too expensive. One way to circumvent that problem is to see the gradient as an expectation and, by sampling our training dataset in small batches (or mini-batches), we can reasonably approximate it. Hence, we use a modified version of the GD called Stochastic Gradient Descend (SGD), which looks like $\nabla_{\pmb{\theta}}\mathcal{L}(\pmb{\theta}; {\bf x}, {\bf y})= \frac{1}{n} \nabla_{\pmb{\theta}} \sum_{i=1}^{n} L_i(\pmb{\theta}; {\bf x}_i, {\bf y}_i)$, where $n$ is the size of a mini-batch (small sample uniformly drawn from the training data set), and $L_i$ is the loss per training example~\cite{ruder:2016,goodfellow:2016,bishop:2024}.

\begin{example}
Consider a simple example to illustrate how gradient descent works. You want to find $\pmb{\theta}^{\ast} = (\theta_1^{\ast}, \theta_2^{\ast})$ that minimizes $f(\pmb{\theta}) = \theta_1^2 + \theta_2^2$.
Next, based on Equation~\eqref{eq:sgd}, we compute the gradient $\nabla_{\pmb{x}} f(\pmb{\theta}) = [2 \theta_1, 2 \theta_2]$. We define an initial guess, $\pmb{\theta}_0 = [1, 0.5]$. From this point, we iteratively apply Equation~\eqref{eq:sgd} for $1{,}000$ iterations with a step size $\eta = 0.01$: 
    \begin{align}
        \pmb{\theta}_{1} &= {\pmb \theta}_0 - \eta \nabla_{\pmb{\theta}} f(\pmb{\theta}_0) \nonumber \\
                         &= [1, 0.5] - 0.01 \times [2 \times 1, 2 \times 0.5] = [0.98, 0.49] \nonumber \\
        \pmb{\theta}_{2} &= {\pmb \theta}_1 - \eta \nabla_{\pmb{\theta}} f(\pmb{\theta}_1) \nonumber \\ 
        &\shortvdotswithin{=} \notag \\
        \pmb{\theta}_{1000} &= {\pmb \theta}_{999} - \eta \nabla_{\pmb{\theta}} f(\pmb{\theta}_{999}) = [0, 0].       \nonumber        
    \end{align}
    After completing the last iteration, this process yields $\pmb{\theta}^{\ast}=[0, 0]$ and we have found the minimum of $f(\pmb{\theta}^{\ast}) = 0$.  
\end{example}

\subsection{Backpropagation}
\label{section:backprop}

In the previous paragraph, we introduced the Gradient Descent, and we saw that it requires
the knowledge of the gradients (derivatives) to correctly identify the proper direction
for descending the gradients. Therefore, when we train a neural network and search for the
proper parameters $\pmb{\theta}$, we use a variant of the GD. Though, how can we estimate 
all the necessary derivatives (gradients)? 
The answer to this question is the Backpropagation
algorithm, which is the powerhouse of modern deep learning. Therefore, we can distinguish 
the training process into two phases: (i) First, backpropagation estimates gradients and 
propagates errors from the output layer to the input. (ii) Then, using these gradients, an 
optimization algorithm minimizes a cost function to find nearly optimal network parameters. 

The backpropagation (BP) algorithm consists of three essential steps:
\begin{enumerate}
    \item The forward pass (propagation), where the input ${\bf x}$ to the neural network activates the units at the input layer, and that activation propagates forward through the hidden layers to the output layer.
    \item The backward pass, where the BP uses the neural network layers’ activations, starting from the output layer and moving backwards to the input, evaluates the gradients (partial derivatives) of the loss
function with respect to those activations. The BP computes gradients at this step using the chain rule of calculus. 
\item The final step involves evaluating derivatives with respect to the network parameters, $\pmb{\theta}$, by forming products of activation from the forward pass and gradients from the backward pass.
\end{enumerate}
Algorithm~\ref{alg:backprop}, which follows~\cite{bishop:2024}, tells us how backpropagation works in a more elegant way. 
\begin{algorithm}
   \caption{{\bf Backpropagation.} An algorithmic representation of the backpropagation algorithm. 
               The algorithm requires a set of parameters $\pmb{\theta}$ (${\bf W}$ and ${\bf b}$), a loss function, $\mathcal{L}$, and the activation functions, $\sigma$ of the neural network $\mathcal{F}$. The algorithms returns the loss function derivatives, $\frac{\partial \mathcal{L}}{\partial w_{ji}}$. }
   \label{alg:backprop}
   \begin{algorithmic}
   \FOR{ $j \in $ all hidden and output units}
        \STATE{$a_j \gets \sum_{i}^{} w_{ji} y_i$}
        \STATE{$y_j \gets \sigma(a_j)$}
   \ENDFOR
   \FOR{ $k \in $ all output units }
        \STATE{ $\delta_k \gets \frac{\partial \mathcal{L}}{\partial a_k}$}
   \ENDFOR 
   \FOR{$ j \in $ all hidden units}
       \STATE{$\delta_j \gets \sigma'(a_j) \sum_{k}^{}w_{kj} \delta_{k}$}
       \STATE{$\frac{\partial \mathcal{L}}{\partial w_{ji}} \gets \delta_j y_i$}
   \ENDFOR%
   \end{algorithmic}
\end{algorithm}
Once the backpropagation algorithm has evaluated the derivatives and propagated the error backward, then we can use an optimizer (see~\ref{section:sgd}) to minimize the loss with respect to the network parameters.

\subsection{Automatic Differentiation}
\label{section:ad}

Backpropagation is a computationally demanding algorithm. Therefore, a fast and accurate method for computing and evaluating the gradients is required. A relatively modern approach to solving that problem is automatic differentiation (AD). AD is a collection of methods to compute and evaluate partial derivatives of functions specified by computer programs~\cite{baydin:2018}. AD should not be confused with numerical or symbolic differentiation. AD builds a computational graph of all the operations within a neural network based on primitive (elementary) arithmetic operations and elementary functions (\emph{e.g.}, $\exp$, $\log$, $\cos$). Applying the chain rule repeatedly on the computational graph can compute all the partial derivatives needed by SGD.\@ 
We refer the avid student to~\cite{goodfellow:2016,burkov:2019, bishop:2024} for more details on optimizers, BP, and AD. It’s worth noting at this point that all modern deep learning frameworks implement backpropagation using Automatic Differentiation (AD).

\subsection{Training Loop}
\label{section:training}
Finally, we can put all the pieces together and see how we can train a neural network. Our first step is to determine the type of neural network we want to use and define it, initialize its parameters, and choose an optimization algorithm (or optimizer) and its learning rate. Moreover, we can choose to use a learning rate scheduler. Learning rate 
schedulers are algorithms that based on some criteria and conditions can change the 
learning rate dynamically during learning. Usually, we write the learning rate as a function
of epochs or iterations and the scheduler uses that information to adjust the learning rate.
Pytorch implements a wide variation of learning rage schedulers\footnote{\url{https://docs.pytorch.org/docs/2.12/optim.html\#how-to-adjust-learning-rate}}. Finally, we have to write our loss function (or criterion). 

Then, we are ready to start training our neural network. We feed the data to the neural 
network in discrete time intervals called epochs. During each epoch, usually, the neural
network is exposed to the entire training dataset, which is chunked into mini-batches. 
Mini-batches are also shuffled to introduce  randomness. For each input or mini-batch, 
the network returns an output or a batch of outputs, which we use to evaluate the loss, 
backpropagate errors, minimize the loss, and eventually update the neural network's 
parameters. Algorithm~\ref{alg:learning} summarizes the learning steps.

\begin{algorithm}
   \caption{{\bf Training Loop.} An algorithmic representation of neural
               network's $\mathcal{F}$ training. The algorithm requires the 
               learning rate $\eta$, and the number of epochs.}
   \label{alg:learning}
   \begin{algorithmic}
       \STATE{Initialize $\mathcal{F}(\pmb{\theta})$}
       \STATE{Set Optimizer (\emph{e.g.}, Adam)}
       \STATE{Set Criterion (\emph{e.g.}, MSE)}
       \STATE{Set learning rate scheduler (\emph{e.g.}, MultiStepLR)}
   \FOR{$e \leq \text{epochs}$}
       \STATE{$\hat{{\bf y}} = \mathcal{F}(\pmb{\theta}; {\bf x})$}
       \STATE{$\mathcal{L}(\pmb{\theta}; {\bf x})= $ Criterion$({\bf y}, \hat{\bf{y}})$}
       \STATE{Compute $\nabla_{\pmb{\theta}} \mathcal{L}(\pmb{\theta}; {\bf x})$}
       \STATE{Update parameters $\pmb{\theta}$ using Optimizer}
   \ENDFOR%
   \end{algorithmic}
\end{algorithm}

\subsection{Potential Issues with Training}
\subsubsection{Vanishing \& Exploding Gradients}
\label{section:exploding}

Let’s take a moment and see two basic cases where the training of a neural network can fail. The first case concerns the initialization of parameters, specifically the magnitude of the initial guess, $\pmb{\theta}_0$, we provide to our optimizer. If too high or too low, the gradients can explode or vanish, respectively, due to overly large or small updates. The second factor is the choice of activation functions, which directly influence gradient behavior. For example, functions like $\tanh$ can saturate (reach their maximal values)
and cause gradients to diminish, while unbounded functions like ReLU can cause them to grow excessively until they explode. These issues are known as vanishing and exploding gradients, respectively. 

One way to mitigate the exploding gradients problem is to use gradient clipping. Gradient
clipping bounds gradients based on either a norm ($\mathrm{grad} = \mathrm{grad} \cdot
\min\{ 1, \frac{b}{|| \mathrm{grad} ||_2} \} $, where $b$ is a constant) or a predetermined
value ($\mathrm{grad} = \max\{ -b, \min\{\mathrm{grad}, b \}  \}$). On the other hand,
to address vanishing gradients, we can use activation functions that prevent gradients from being squeezed toward zero, such as the rectified linear unit (ReLU)~\cite{glorot:2011} or
alternatively, recurrent neural networks like LSTM~\cite{hochreiter:1997}.

\begin{example}
Imagine a feed-forward neural network with 7 layers. The signal moves from the input layer (first) to the output layer (last). Recall the backpropagation algorithm (Algorithm~\ref{alg:backprop}): as we propagate the error backward, we multiply the gradient by a number at each layer. Assume each layer uses the same multiplier. If we start with a small value, say w = 0.5, the last layer receives the full signal, but by the time it reaches the first layer (in backward propagation), the value is tiny: $0.5^7$. No actual information reaches the first layer—this is the vanishing gradient. If we choose $w = 3$, the first layer receives a huge value, $3^7$, during backward propagation; we call this the exploding gradient. 
\end{example}

\subsubsection{Generalization Error \& Regularization}
\label{section:bn}

\begin{figure}[!htpb]
    \centering
    \includegraphics[width=0.8\linewidth]{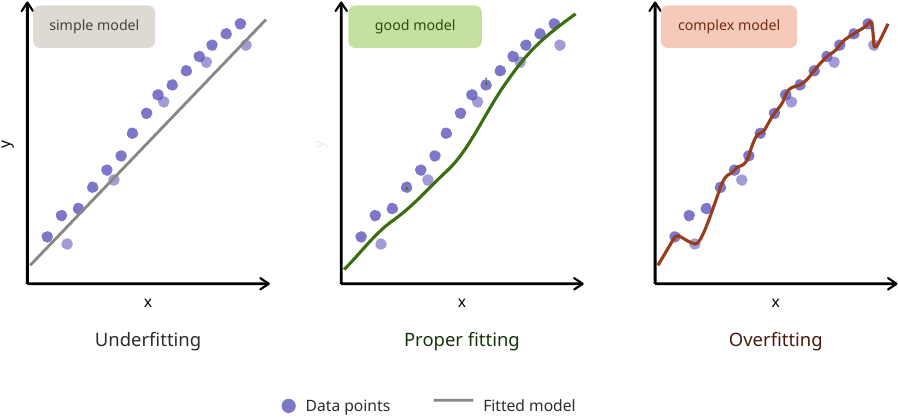}.
    \caption{{\bf Underfitting \& Overfitting.} In this figure we visualize the concepts of 
    underfitting and overfitting. We assume we try to fit a model, colored lines, to our two-dimensional data points, blue nodes. We see that when our model is too simple, we underfit the 
    data (left panel, high bias, the model misses the trend), and in case where our model is too
    complex we overfit the data (right panel, high variance, the model learns the noise). On the other hand, when our model is balanced we can see that fits the data and generalizes well (middle panel).}
    \label{fig:overfitting}
\end{figure}
Another issue we may face when training neural networks is overfitting or underfitting.
First, let's introduce the generalization error. The generalization error, is the error
a neural network commits when operating on data that has not previously
seen~\cite{goodfellow:2016}, provided that the unseen data still follow the same 
distribution as the training data. Therefore, when a neural network performs well even on 
unseen data, we say that it generalizes well. To assess the generalization error,
we split the data set into three disjoint subsets: training, test, and validation data sets.
We train and validate the neural network on the first two sets and evaluate the 
generalization error on the test data set. 

Now, when we have a neural network that cannot \emph{learn} the primary trends and 
statistics of the data and exhibits significant generalization error, we say the network is 
underfitting the data (i.e., the neural network is too simple). On the other hand, we may 
end up with a neural network that \emph{learns} the training data too well and fails to 
generalize~\cite{burkov:2019} (complex neural network). In this case, we say that the neural 
network is overfitting the data. We visualize the concepts of underfitting and overfitting 
in Figure~\ref{fig:overfitting}. 

Thus, the question: ``How can we prevent a neural network from overfitting or underfitting?’’ emerges. So, one way to prevent overfitting or underfitting is to use a regularization method. And what is a regularization? Regularization refers to machine learning methods that reduce generalization errors (i.e., on the test set). Two of the most common regularization techniques are $L_1$ and $L_2$, which add a penalty term to the loss function. This term plays a unique role depending on whether
it's $L_1$ or $L_2$. In $L_1$ regularization, the penalty term makes the neural network (model) sparse, meaning most of its parameters are zero. On the other hand, in $L_2$ regularization, the penalty term pushes the model's weights toward small values near zero, preventing overfitting and keeping the model simple. There are more regularizers that 
can prevent overfitting, such as Dropout and Batch Normalization. Dropout randomly ``kills'' neurons (units) within a neural network's layer (or layers)~\cite{srivastava:2014}, and
Batch Normalization recenters and rescales the input to a neural network layer. Again,
we refer the students to Chapter $7$ of~\cite{goodfellow:2016} for more information about regularization methods.

Batch normalization is an algorithm that receives a mini-batch as input, computes its mean 
and variance, and then applies a normalization step followed by a scale-and-shift 
transform~\cite{ioffe:2015}. By doing so, it can stabilize learning and even speed it up~\cite{ioffe:2015}. Since we are interested in solving partial differential equations,
we want to know if Batch Normalization can help us improve our algorithm's performance. 
Indeed, researchers have used batch normalization to accelerate learning when solving differential equations with deep neural networks~\cite{blechschmidt:2021}.  

Let $\mathcal{B} = \{ {\bf x}^1, {\bf x}^2, \ldots, {\bf x}^n \}$ be a mini-batch
of size $n$ with ${\bf x}^i = ({\bf x}^i_1, \ldots, {\bf x}^i_d)$. The batch
normalization transform is given by
\begin{align}
    \label{eq:bn}
    \begin{split}
        \hat{{\bf x}}^i_j &= \frac{{\bf x}^i_j - \mu^i_{\mathcal{B}}}{\sqrt{(\sigma^i)^2_{\mathcal{B}} + \epsilon}}, \quad i\in [1, n], \, j\in [1, d], \\
        \hat{{\bf y}}^{i,l}_j &= \alpha^i \hat{{\bf x}}^i_j + \beta^i, \quad l \in [1, L].
    \end{split}
\end{align}
where $\mu^i_{\mathcal{B}}$ and $\sigma^i_{\mathcal{B}}$ are the per-dimension
mean and variance, respectively. $\epsilon$ is a tiny constant added to the
denominator for numerical stability, and $\alpha$ and $\beta$ are learnable
parameters. $\hat{\bf y}^{i,l}$ here is the transformation output, and it
refers to the output of a layer $l$. Equation~\eqref{eq:bn} is used in training
only. The population statistics replace the mean and the variance during the
inference. The reader can find more details about BN in~\cite{ioffe:2015}.

\section{Universal Approximation Theorem}
\label{section:universal}

\begin{figure}
    \centering
    \includegraphics[width=0.45\textwidth]{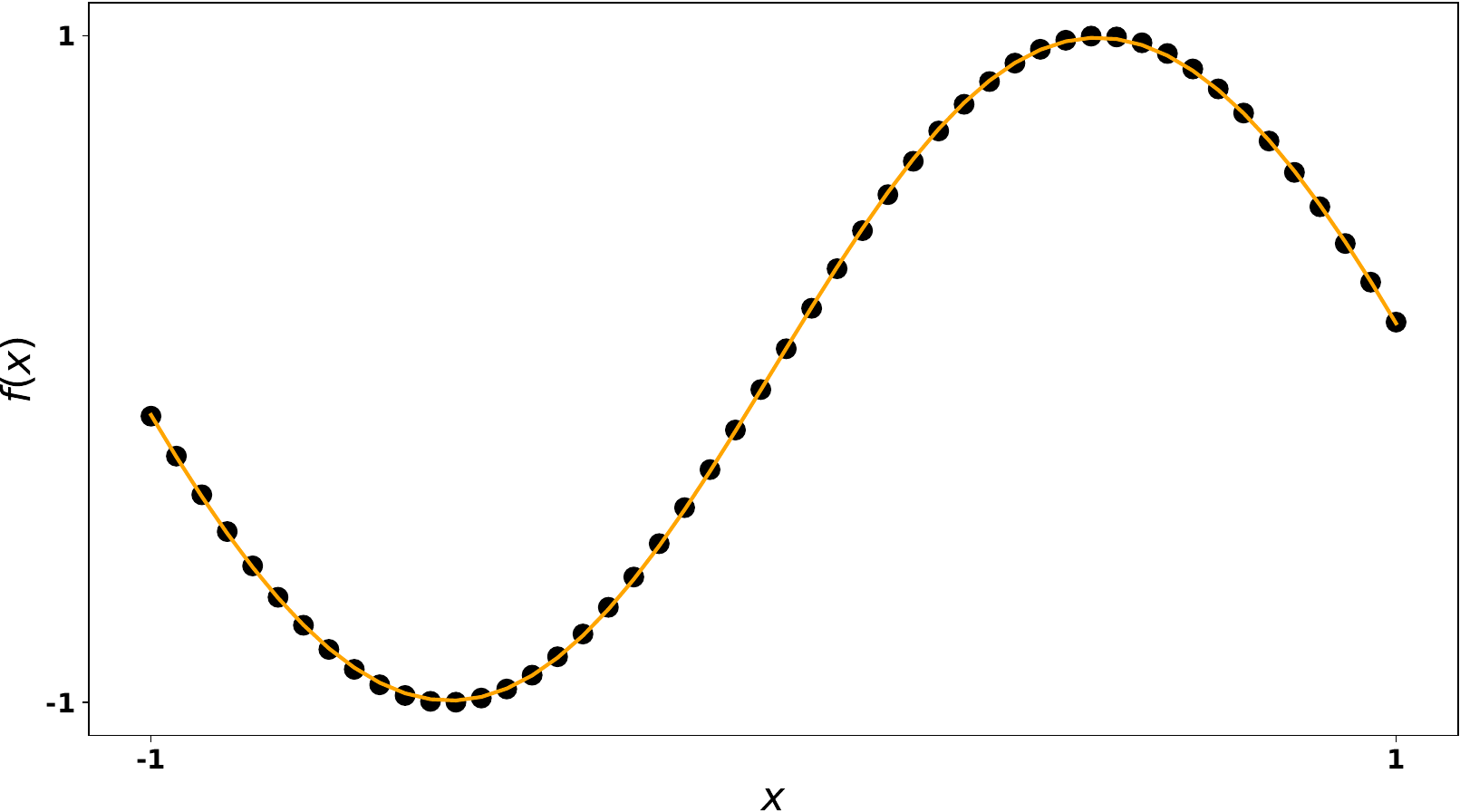}
    \caption{{\bfseries \sffamily Universal Approximation Theorem.} A neural
    network with one hidden layer, three hidden units, and a $\tanh$ activation
    function approximates the function $f(x)= \sin(3x)$ in the interval $[-1,
    1]$. Fifty samples of the function $f(x)$ used to train the neural network
    are shown here as black dots. The orange line indicates the approximation,
    $\hat{f}(x)$, provided by the neural network.}
    \label{fig:uat}
\end{figure}

In the previous paragraphs we introduced the ideas of neural networks, how to train 
neural networks, and how to regularize them to avoid some of the pitfalls of learning. 
Therefore, we are almost ready to move forward and see how we use deep neural networks
to approximate the solutions of partial differential equations. But before we dive into
all those details we have to make sure that our neural networks can actually operate 
as function approximators. 

Now, recall the feed-forward neural network or MLP we introduced in 
Section~\ref{section:mlp}.
One of the well-established and fundamental properties of a feed-forward neural network is that of a universal approximator, meaning that we can use an MLP to approximate any function
given enough hidden units in our hidden layer. 

\begin{example}
    For instance, let’s assume we would like to approximate the function $f(x) = \sin(3x)$ in the interval $[-1, 1]$. We can use a simple MLP with one input, one output, one hidden layer, and a $tanh$ activation function applied on its hidden units, 
\begin{center}
    \FINL(1, 3) $\leadsto$ Tanh $\leadsto$ \FHL(3, 3) $\leadsto$ Tanh $\leadsto$ \FOL(3, 1),
\end{center}
to approximate the function $f$. In Figure~\ref{fig:uat}, we see in black dots the function $f(x) = \sin(3x)$, and in orange (line) the approximated, by the neural network, function $\hat{f}(x)$, after training the network on the data represented by the black dots.
\end{example}

If we are to put the universal approximation theorem in a more mathematical context, we should introduce theorem proposed in~\cite{cybenko:1989,hornik:1989}.
Therefore, let $I_q$ be a $q$-dimensional unit cube, $[0, 1]^q$, and $C(I_q)$ the space of continuous functions on $I_q$, and $\alpha_j, \beta_j \in \mathbb{R}$ fixed parameters.
\begin{theorem} Let $\sigma$ be any continuous sigmoidal function. Then, the
    finite sums of the form \begin{align} G(\X) &= \sum_{j=1}^{M} \alpha_j
    \sigma({\bf y}_j^T \X + \beta_j), \end{align} are dense in $C(I_n)$. In
    other words, given any $f\in C(I_n)$ and $\epsilon > 0$, there is a sum,
    $G(\X)$ of the above form, for which \begin{align} |G(\X) - f(\X)| <
    \epsilon \quad \text{ for all } \X \in I_n.
    \end{align}
\end{theorem}
In simple words, a feed-forward neural network with a single hidden layer followed by a nonlinear activation function such as a sigmoid ($\sigma(x) = \frac{1}{1 + \exp(-x)}$) and a linear output layer can approximate a continuous function defined on a compact subset of $\mathbb{R}^q$. Although the theorem states that it is possible to approximate any such function, it becomes apparent in practice that we need to discuss the number of hidden units we will have to use to approximate a function of interest. However, deep feed-forward neural networks seem to work in practice and approximate functions defined on compact subsets of $\mathbb{R}^q$. Furthermore, two variations of the Universal Approximation theorem (UAT)
have been proposed: The first one by Leshno et al.~\cite{leshno:1993} on non-polynomial functions, including the ReLU functions, and the second by Hornik et al.~\cite{hornik:1990} discussing the first derivatives of continuous functions.

\section{Differential Equations \& Deep Learning}

In the previous part of this article, we visited the notions of neural networks. We saw how to train a neural network, potential training issues, and methods to mitigate them. We also provided the necessary algorithms and PyTorch implementations. Now, we will set aside deep learning and neural networks. We will focus on aspects of differential equations to help the reader make the transition from neural networks and deep learning to partial differential equations and finally how to combine them.

\subsection{Partial Differential Equations}

Partial differential equations (PDEs) typically involve partial derivatives of the dependent
variable with respect to time and space, and they describe distributed physical systems 
(compared to lumped systems described by ordinary differential equations, where the 
dependent variables are functions of time alone).

In general, we define a PDE as a system of order $m \in \mathbb{N}$:
\begin{align}
    \label{eq:pde}
    {\bf F}(\X, D\Y, D^2\Y, \ldots, D^m\Y) &= 0,
\end{align}
where $\X \in \Omega \subseteq \mathbb{R}^q, \, q \geq 2$ are the independent
variables, and $\Y = (y_1, y_2, \ldots, y_p)$ are the unknown functions (or dependent).
All the functions $y_i, \, i=1,\ldots,p$ can be functions of space and time. We can write
all the partial derivatives as 
\begin{align}
  D &= \frac{\partial^{\alpha}y}{\partial x_1^{\alpha_1} \partial x_2^{\alpha_2} \cdots \partial x_q^{\alpha q}}.   
\end{align}
When $p \geq 2$, then we have a system of PDEs; otherwise, we will say that we have a PDE or a scalar PDE.
\begin{figure}
\begin{minipage}{.4\textwidth}
    \centering
    \includegraphics[width=0.4\textwidth]{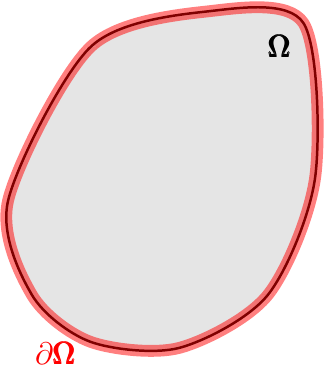}
\end{minipage}
\hfill
\begin{minipage}{.6\textwidth}
    \centering
    \includegraphics[width=1.\textwidth]{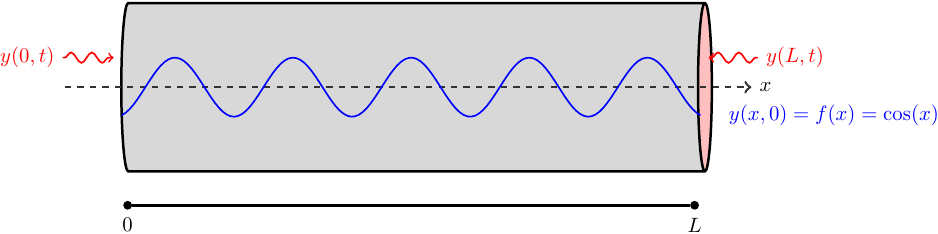}
\end{minipage}
\caption{{\bfseries \sffamily Domain and boundary schematics.} Left Panel shows
    a domain $\Omega$ of a partial differential equation in gray color and its
    boundary $\partial \Omega$ in red color. The right panel shows a rod of
    size $L$ (gray color), which is used to simulate heat diffusion (see main
    text). The boundary conditions $y(x, 0)$ and $y(x, L)$ are depicted with
    red color fonts, and the initial condition $y(x, 0) = f(x) = \cos(x)$ with
    a blue sinusoidal line.}
    \label{fig:domain-rod}
\end{figure}

We usually study PDEs as initial-boundary-value-problems (IBVP), where we
define a domain, $\Omega$, a boundary $\partial \Omega$, and some initial
conditions when $t=0$. Boundary Conditions are, in essence, constraints imposed
on the boundary of a PDE's domain, and the solution of a PDE must satisfy these
boundary and initial conditions. Example~\ref{example:3} shows how we write a 
PDE (one-dimensional heat equation) with initial and boundary conditions.
In addition, we see in the left panel of Figure~\ref{fig:domain-rod}
a schematic representation of a domain $\Omega$ in gray color and its 
boundary $\partial \Omega$ in red. 

\begin{example}
\label{example:3}
In this example, we see the partial differential equation of the one-dimensional heat equation along with some initial and boundary conditions.
    \begin{align}
        \frac{\partial y(x, t)}{\partial t} &= \alpha \frac{\partial^2 y(x, t)}{\partial x^2} && \text{PDE} \nonumber  \\
        y(x, 0) &= \sin(\pi x) \qquad \text{for } 0 < x < 1, && \text{Initial Conditions} \nonumber \\
        y(0, t) &= 0 \qquad \text{for } t > 0,  && \text{Boundary Condition}  \nonumber \\
        y(1, t) &= 0 \qquad \text{for } t > 0. && \text{Boundary Condition} \nonumber
    \end{align}
\end{example}

Now, to better understand the notions of boundaries and initial conditions, let's 
imagine a rod for which we want to study heat diffusion when we apply a heat
source on one of its ends at time $t=0$. The inside of the rod is the domain,
$\Omega$, where we define and solve the heat equation to obtain the diffusion
inside the rod (see the gray area in the right panel of Figure~\ref{fig:domain-rod}).
On the other hand, the boundary is the surface of the rod and its left and right
edges (see the left panel of Figure~\ref{fig:domain-rod}, red color again).
On the boundary, we can assume that different things are happening. For example,
suppose we place the heat source on the right side of the rod and assume the rest of the rod is
insulated. In that case, we will impose a boundary condition on the right side
that depends on time and provides the necessary heat. Moreover, we can assume
that the heat equation holds on the remaining part of the boundary, with
constant zero boundary conditions. Therefore, we see that boundary conditions
tell us how the heat equation behaves on the surface of the cylinder and at its
ends. Students who would like to learn more on PDEs are encouraged to read~\cite{strauss:2007,farlow:1993}.   

\begin{tcolorbox}[title={\bf Boundary Conditions}]
    We typically use five basic types of boundary conditions (or a mixture of those):
\begin{itemize}
    \item {\bf Dirichlet} is the simplest boundary condition (BC), where $y({\bf x}, t) = g({\bf x}, t)$,
    when ${\bf x} \in \partial \Omega$. We use this type of BC when a physical quantity, such as temperature,
    is kept constant on the boundary.  
    \item {\bf Neumann} is described by $\frac{\partial y({\bf x}, t)}{\partial {\bf n}} = g({\bf x}, t)$,
    where ${\bf n}$ is the (exterior) normal to the boundary, and $\frac{\partial y({\bf x}, t)}{\partial {\bf n}}$
    is the normal derivative. We use the Neumann boundary conditions when there is a reflection,
    a flux, or a gradient at the boundary. 
    \item {\bf Robin} follows the equation $a y({\bf x}, t) + b \frac{\partial y({\bf x}, t)}{\partial {\bf n}} = g({\bf x}, t)$,
    where $a,b\in\mathbb{R}$ constants, and ${\bf n}$ is the normal vector on the boundary.
    \item {\bf Mixed} conditions are combinations of any of the previous types. For instance, 
    on one end of a rod, we might have constant temperature, thus a Dirichlet boundary condition,
    and on the other end, a flux of heat, hence a Neumann boundary condition. 
    \item {\bf Cauchy} guarantees the uniqueness of a solution by imposing a Dirichlet and a
    Neumann boundary condition simultaneously. 
\end{itemize}
\end{tcolorbox}
\noindent\begin{minipage}{\textwidth}
\label{box:boundaryconds}
\end{minipage}

\subsection{Numerical Solutions of PDEs}

Let's return to Equation~\eqref{eq:pde} and rewrite it as an
initial-boundary-value-problem (\BV) of an unknown spatiotemporal function
$y({\bf x}, t)$ defined on the domain $\Omega \times [0, T]$, where $\Omega \in
\mathbb{R}^d$, and $\partial \Omega$ is the boundary of $\Omega$. Then $y$
satisfies the IBVP:
\begin{align}
    \label{eq:partial}
    \begin{split}
        (\partial_t + \mathcal{T})y({\bf x}, t) & = 0,  \phantom{aaaaa} ({\bf x}, t)\in \Omega \times [0, T] \\
        y({\bf x}, 0) &= y_0({\bf x}), \phantom{aa} {\bf x} \in \Omega \\
        y({\bf x}, t) &= g({\bf x}, t), \phantom{a} ({\bf x}, t) \in \partial \Omega \times [0, T],
    \end{split}
\end{align}
where $\mathcal{T}$ is an operator that describes the PDE, for instance, when
$\mathcal{T}$ is a linear operator, then the PDE described by
Equation~\eqref{eq:partial} is a linear PDE. The function $y_0(\X)$ represents
the initial conditions, and the function $g(\X, t)$ indicates the type of
boundary conditions, such as Dirichlet or Neumann (see Box~\ref{box:boundaryconds}).

\begin{figure}
    \centering
    \includegraphics[width=0.45\textwidth]{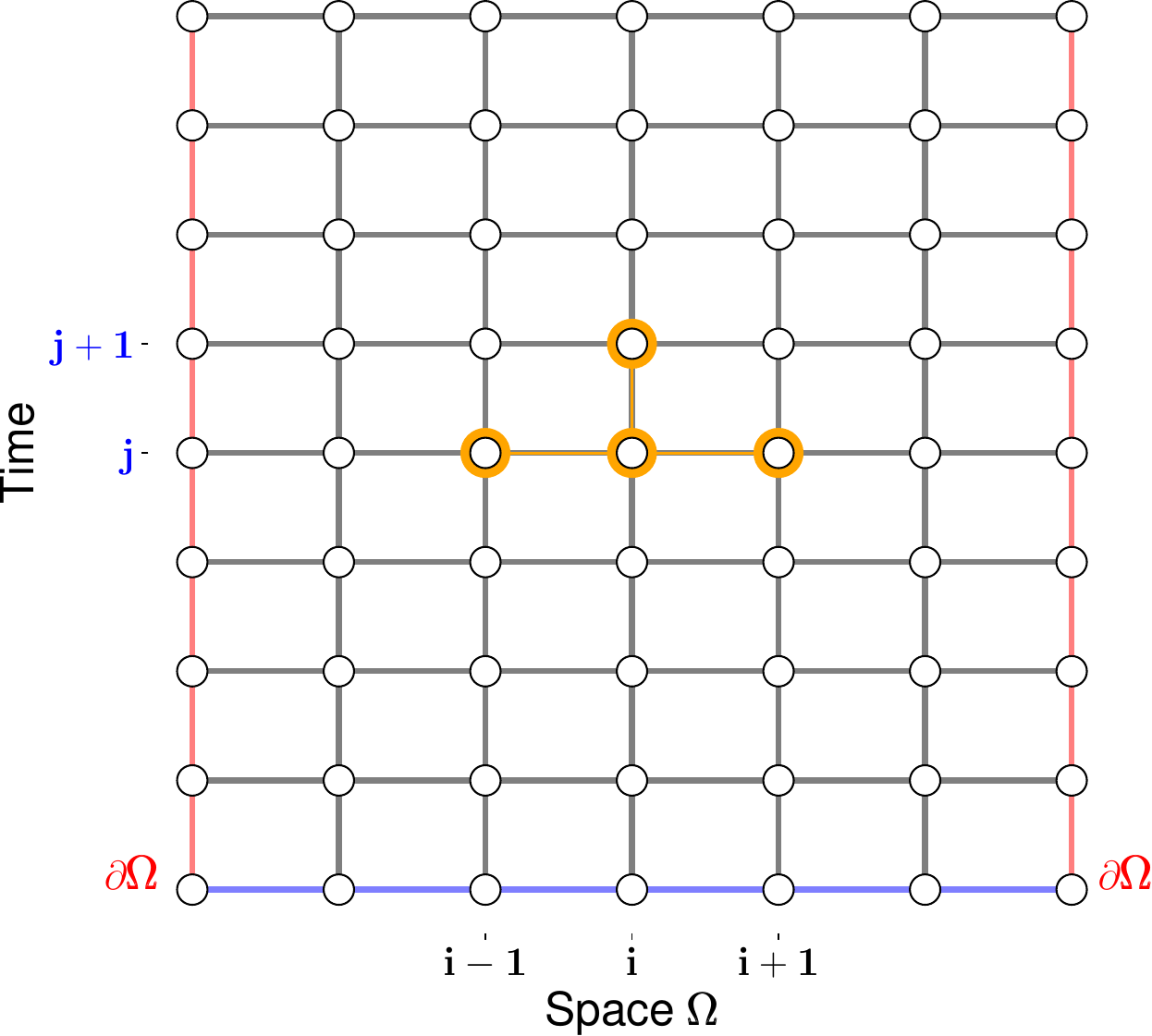}
    \caption{{\bfseries \sffamily One-dimensional Finite Differences Scheme}
    for a one-dimensional problem. The spatio-temporal discrete grid appears as
    circles and and the stencil of the finite differences is shown in orange
    color. The blue line indicates the initial conditions, and the red one
    shows the boundary conditions. The blue $j$s and black $i$s reflect the
    temporal and spatial discrete steps, respectively. }
    \label{fig:finite_diffs}
\end{figure}

In this article, we will forget about analytical solutions or theoretical approaches to 
PDEs, and we will focus on numerical solutions. To this end, one way to numerically solve
an \BV given by Equation~\eqref{eq:partial} is to use the finite differences (FD) method. 
The main idea behind the FD methods is to discretize the domain of 
Equation~\eqref{eq:partial} using regular grids, both temporal and spatial, as shown in 
Figure~\ref{fig:finite_diffs}. At the spatial nodes of the grid, the FD will estimate 
spatial derivatives; at the temporal nodes, time derivatives (see Box~\ref{box:fd}
for a brief introduction to finite differences).
Besides the Finite Differences method, other, more accurate numerical methods are used to 
solve PDE problems.
One such method is the Galerkin, which uses linear combinations of basis functions to 
approximate the solution of a PDE.\@ 

Let's formulate the general idea behind Galerkin methods through an example. Take the one-dimensional
Poisson equation:
\begin{align}
    \label{eq:poisson}
    \frac{\partial^2 y}{\partial x^2} &= f, \quad \text{on } \Omega, \nonumber \\
    y &= 0, \quad \text{on } \partial \Omega,
\end{align}
and we want to find a numerical approximation $\hat{y}$. To do so, we first
define a finite-dimensional approximation space $\mathcal{X}_0^K$ and an
associated set of basis functions $\{\phi_i \in \mathcal{X}_0^K \}$, where
$i=1,\ldots, k$. In addition, we require the basis functions to satisfy the
boundary conditions such that $\phi_i = 0$ on $\partial \Omega$. We want to
find a numerical approximation $\hat{y}(\X) = \sum_{j=1}^{k} \hat{y}_j
\phi_j(\X)$ that satisfies
\begin{align}
    \label{eq:galerkin}
    -\int_{\Omega} \nabla v \cdot \nabla \hat{y}\, dV &= \int_{\Omega} v f dV, \quad \forall v \in \mathcal{X}_0^K,
\end{align}
where $v \in \mathcal{X}_o^N$ is a test function.  
The problem described by Equation~\eqref{eq:galerkin} is a variational problem, where we are looking to optimize
functionals such that we obtain a numerical approximation, $\hat{y}$, of the solution $y$ of Equation~\eqref{eq:poisson}.
Equation~\eqref{eq:galerkin} constitutes the Galerkin formulation, and from that equation, one can derive
the Finite Elements, Spectral Elements, and Spectral Formulation methods. The details of those methods are beyond
the scope of this article; therefore, we will omit them. However, students who are interested in this
topic can find more details in~\cite{langtangen:2019, blanchard:2012}.

\begin{tcolorbox}[title={\bf Finite Differences}]
\label{box:1}
    Let the differential operator $\mathcal{T}$ be $-\frac{\partial^2 }{\partial x^2}$ and take as boundary conditions $y(0, t) = y(1, t) = 0$, and
    initial conditions $y(x, 0) = y_0(x)$, then we can rewrite Equation~\eqref{eq:partial} in one-dimension as:
    \begin{align}
        \label{eq:fd_example}
        \begin{split}
             \frac{\partial y(x, t)}{\partial t} &= \frac{\partial^2 y(x, t)}{\partial x^2}, \\
             y(x, 0) &= y_0(x), \\
             y(0, t) &= y(1, t) = 0.
        \end{split} 
    \end{align}
Then, we can numerically solve the problem given by
Equation~\eqref{eq:fd_example} by discretizing the temporal and spatial
derivatives using finite differences. One such method is the forward explicit
scheme, where we discretize the temporal and spatial derivatives using $M \in \mathbb{N}$
discrete nodes for the time derivatives and $N \in \mathbb{N}$ for the spatial. Hence, the
temporal and spatial nodes are given by $t_j = j \Delta_t$, where $j = 0,
\ldots, M$ and $x_i = i \Delta x$, where $i = 0, \ldots, N$, respectively. Then
we obtain the grid points (or nodes) $(x_i, t_j)$ on which we will approximate
the solution $y(x_i,t_j)$. Figure~\ref{fig:finite_diffs} shows the
spatio-temporal discrete grid and the stencil (orange color) of the finite
differences we use in this example. The blue color indicates initial
conditions, and the red one shows the boundary conditions.  Finally, we
approximate the temporal derivative using a forward difference approximation
and the spatial derivative with a central difference approximation:
\begin{align}
    \label{eq:finiti_diffs}
        \frac{\partial y(x, t)}{\partial t} &\approx \frac{y_{i, j+1} - y_{i, j}}{\Delta_t}, & 
        \frac{\partial^2 y(x, t)}{\partial x^2} &\approx \frac{y_{i+1, j} - 2 y_{i, j} + y_{i-1, j}}{\Delta_x^2}.
\end{align}
Now, we plug the approximated derivatives of Equation~\eqref{eq:finiti_diffs} to Equation~\eqref{eq:fd_example}
and we obtain
\begin{align}
    \label{eq:fd_solution}
    y_{i, j+1} &= y_{i, j} + \alpha \Big( y_{i+1, j} - 2 y_{i, j} + y_{i-1, j} \Big),
\end{align}
where $\alpha = \frac{\Delta t}{\Delta x^2}$. We apply a similar discretization
on the boundary conditions, and we can then either iterate
Equation~\eqref{eq:fd_solution} over $i$ and $j$ and obtain the solution, or we
can rewrite Equation~\eqref{eq:fd_solution} as a system of linear equations and
solve it. Avid students can find more information about finite difference methods 
in~\cite{morton:2005, strikwerda:2004}. 
\end{tcolorbox}
\noindent\begin{minipage}{\textwidth}
\label{box:fd}
\end{minipage}


\subsection{Numerical Solutions of PDEs with Deep Galerkin Method}
\label{section:dgm}

\begin{figure*}
    \centering
    \includegraphics[width=0.7\textwidth]{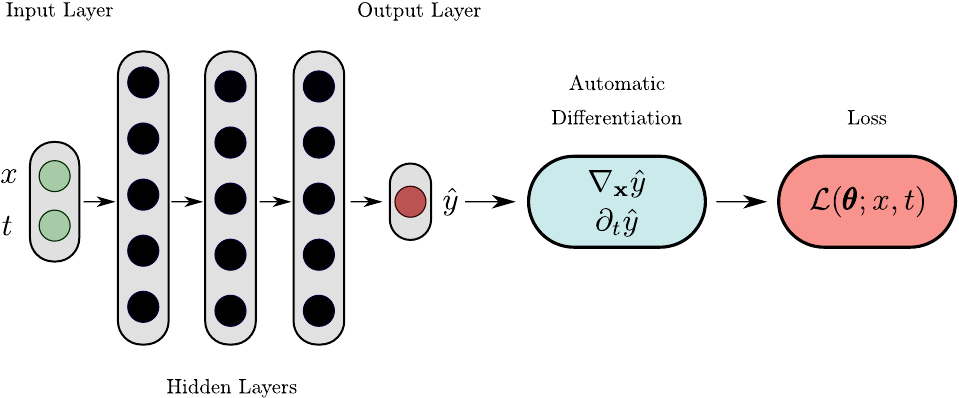}
    \caption{{\bfseries \sffamily Deep Galerkin Method Schematic}. A feed-forward neural network (MLP) 
    takes an input $(x, t)$ (here we assume that the $x\in\mathbb{R}$) and returns an output $\hat{y}$.
    Then, we use automatic differentiation to compute the derivatives $\partial_t \hat{y}$ and any spatial
    derivative $\nabla_x \hat{y}$. Then, we plug the derivatives in the loss function~\eqref{eq:dgm_loss}
    and use the result to train the neural network. See main text and algorithm~\ref{alg:dgm} for more
    details on the DG method.}
    \label{fig:dgm_cartoon}
\end{figure*}

Now, we will see how we combine the Galerkin method with deep learning to approximate solutions of PDEs. 
The deep Galerkin (DG) method, introduced by Sirignano and Spiliopoulos in~\cite{sirignano:2018}, 
extends the classical Galerkin method. Its main idea is to define a loss function based on an initial-boundary-value problem,
minimize it by sampling points from the domain, the boundary, and the initial conditions, and use a
neural network to approximate the solution. 

Therefore, our first step is to determine our loss function. The loss function should take into account
all the available information (initial conditions and boundary conditions) about our problem and combine
it with the outputs of a neural network, $\mathcal{F}$. Then an optimizer will try to minimize our loss
and find the proper parameters, $\pmb{\theta}$, of a neural network that sufficiently approximates the
solution to our PDE. Thus, we define our loss function as:
\begin{align}
    \label{eq:dgm_objective}
    \begin{split}
            \mathcal{J}(\mathcal{F}({\bf x}, t; \pmb{\theta})) &= \underbrace{|| (\partial_t + \mathcal{T})\mathcal{F}({\bf x}, t; \pmb{\theta}) ||_{[0, T]\times \Omega, \nu_1}^{2}}_{\text{Differential Operator}} \\
            &+ \underbrace{||\mathcal{F}({\bf x}, t; \pmb{\theta}) - g({\bf x}, t) ||_{[0, T] \times \partial \Omega, \nu_2}^{2}}_{\text{Boundary Conditions}} \\
            &+ \underbrace{|| \mathcal{F}({\bf x}, 0; \pmb{\theta}) - y_0({\bf x}) ||_{\Omega, \nu_3}^{2}}_{\text{Initial Conditions}},
    \end{split}
\end{align}
where the first term tells us how good our approximation of the differential operator is, and the second and third terms indicate how well we approximate the boundary and the initial values, respectively. The norm is an $L^2$ given by $||f(y)||_{\mathcal{Y}, \nu}^{2} = \int_{\mathcal{Y}} |f(y)|^2 \nu(y) dy $
with $\nu$ being a density defined on $\mathcal{Y}$ ( $\mathcal{Y}$ becomes $[0, T]\times \Omega$ for the PDE, $[0, T] \times \partial \Omega$ for the boundary conditions, and $\Omega$ for the initial conditions). Once, we have trained our neural networks and thus the loss has been minimized, $\mathcal{J}(\mathcal{F}(\pmb{\theta}; {\bf x}, t)) \rightarrow 0$, then the output 
of our neural network, $\mathcal{F}(\pmb{\theta}; {\bf x}, t)$ is the approximation, $\hat{y}$, of our \BV (Equation~\eqref{eq:partial}). 

A potential problem with our method is that, as the dimension of the domain and the variables $q$ increase, integrals over
our domain $\Omega$ become intractable. To address this, we sample from distributions on $\Omega$ and $\partial \Omega$ to estimate the squared error:
\begin{align}
    \label{eq:dgm_loss}
    \begin{split}
     \mathcal{L}(\pmb{\theta}) &= \Big( \partial_t \mathcal{F}({\bf x}_{\Omega}, t_{\Omega}; \pmb{\theta}) + \mathcal{T} \mathcal{F}({\bf x}, t_{\Omega}; \pmb{\theta}) \Big)^2 \\
            &+ \Big( \mathcal{F}({\bf x}_{\partial \Omega}, t_{\partial \Omega}; \pmb{\theta}) - g({\bf x}_{\partial \Omega}, t_{\partial \Omega}) \Big)^2 \\
            &+ \Big( \mathcal{F}({\bf x}_{t_0}, 0; \pmb{\theta}) - y_0({\bf x}_{t_0}) \Big)^2,
    \end{split}
\end{align}
where $({\bf x}_{\Omega}, t_{\Omega}) \sim \mathcal{U}(\Omega \times [0, T])$,
$({\bf x}_{\partial \Omega}, t_{\partial \Omega}) \sim \mathcal{U}(\partial
\Omega \times [0, T])$, and ${\bf x}_{t_0} \sim \mathcal{U}(\Omega)$. By doing so, we have to find
the proper parameters $\pmb{\theta}$ that minimize Equation~\eqref{eq:dgm_loss}, instead of
Equation~\eqref{eq:dgm_objective}, to find an approximated solution to problem~\eqref{eq:partial}.

In Figure~\ref{fig:dgm_cartoon} we see a schematic of the main idea behind DG. 
We see how a neural network (left side of the picture) receives as input (olive green nodes), $\X$,
the independent variables (\emph{i.e.}, time, and spatial components) of the \BV, and the output 
layer (red node) returns the approximation to the solution, $\hat{y}$. Automatic differentiation uses
the inputs and the output of the neural network to estimate the temporal
(\emph{i.e.}, $\partial_t \hat{y}$) and the spatial (\emph{i.e.}, $\nabla_{\bf x}\hat{y}$) derivatives
(teal colored box). More precisely, we use the \texttt{autograd}\footnote{\url{https://docs.pytorch.org/tutorials/beginner/blitz/autograd\_tutorial.html}} package of Pytorch to estimate the derivatives. Finally, we
estimate the loss function and proceed with its minimization (red box). 

\begin{algorithm}
    \caption{{\bf Deep Galerkin Method.} An algorithmic representation of the DG method. 
    The algorithm requires the number of iterations \texttt{n\_iters}, a neural network, $\mathcal{F}(\pmb{\theta})$,
    an optimizer, and a learning rate $\eta$.}
    \label{alg:dgm}
    \begin{algorithmic}
        \STATE{Initialize $\mathcal{F}(\pmb{\theta})$}
        \STATE{Initialize an optimizer}
    \FOR{$i \leq $n\_iters}
     \STATE{Generate $(\X, t)_{\Omega \times [0, T]} \sim \nu_{1}$}
     \STATE{Generate $(\X, t)_{\partial \Omega \times [0, T]} \sim \nu_{2}$}
     \STATE{Generate $(\X, 0)_{\Omega} \sim \nu_3$}
    
     \STATE{Compute $ y_{\Omega \times [0, T]} = \mathcal{F}(\pmb{\theta};(\X, t)_{\Omega \times [0, T]})$}
     \STATE{Compute $ y_{\partial \Omega \times [0, T]} = \mathcal{F}(\pmb{\theta};(\X, t)_{\partial \Omega \times [0, T]})$}
     \STATE{Compute $ y_{\Omega, t=0} = \mathcal{F}(\pmb{\theta};(\X, 0)_{\Omega})$}
     \STATE{Compute $\partial_t \hat{y}$ and $\nabla_{\X}\hat{y}$}
     \STATE{Compute $\mathcal{L}(\pmb{\theta}; {\bf x}, t)$ based on equation~\eqref{eq:dgm_loss}}
     \STATE{Compute $\Delta \pmb{\theta} = \eta \nabla_{\theta} \mathcal{L}(\pmb{\theta};\X, t)$}
     \STATE{Update parameters $\pmb{\theta}$ using the optimizer}
     
     \IF{\text{learnig rate schedule condition met}}
     	\STATE{Adjust $\eta$}
     \ENDIF%
    \ENDFOR%
    \end{algorithmic}
\end{algorithm}
Algorithm~\ref{alg:dgm} outlines the steps to minimize the loss function~\eqref{eq:dgm_loss}. First, initialize the neural network's parameters $\pmb{\theta}$ (weights and biases). Next, set the number of learning iterations and select an optimizer. For each iteration, randomly draw samples from the distributions $\nu_1$, $\nu_2$, and $\nu_3$, corresponding to the domain $\Omega$, boundary $\partial \Omega$, and initial conditions. Feed these samples into the neural network to compute $y_{\Omega \times [0, T]}$, $y_{\partial \Omega \times [0, T]}$, and $y_{\Omega, t=0}$, for the domain, boundary, and initial time $t=0$. Compute the loss using Equation~\eqref{eq:dgm_loss}, backpropagate the errors, and update $\pmb{\theta}$ with the optimizer. If using a learning rate scheduler, check whether its conditions are satisfied, and update the learning rate $\eta$ accordingly. With a PyTorch scheduler, manual checking is unnecessary, as it manages conditions automatically.

\begin{tcolorbox}[title={\bf Absolute Error}]

There are many ways to evaluate numerical solutions of PDEs. In this article, 
we use one of the simplest and easiest to implement, the absolute error 
between the approximated by a neural network solution, \(\hat{{\bf y}}\), and 
the exact solution, \({\bf y}\). 
Moreover, because we use mini-batches to train our neural network
we resort to the \emph{mean absolute error} (MAE), so that we average the errors over batches. 
\begin{align}
\label{eq:mae}
 \text{MAE}({\bf y}, \hat{{\bf y}}) &= \frac{1}{n} \sum_{i=1}^{n} |y_i - \hat{y}_i|.
\end{align}
To numerically evaluate the MAE we use the function \texttt{mean\_absolute\_error} 
provided by the \texttt{Scikit-learn} Python package~\cite{scikit-learn}.
Bear in mind, when we use the deep Galerkin method for real problems, we may not know
the exact solution. In that case, we can measure the error from a solution produced by
a finite element method.
\end{tcolorbox}
\noindent\begin{minipage}{\textwidth}
\label{box:mae}
\end{minipage}

\section{A Complete Example: The 1D Heat Equation}
\label{section:heat_example}

Now, it's time to piece everything together and see how we can use the deep Galerkin method to solve
a partial differential equation. For simplicity, we will use the one-dimensional heat equation with
appropriate initial and boundary conditions. Next, we will demonstrate how to derive the loss function
and implement it in PyTorch. We will then address considerations such as mini-batch size, batch normalization,
and, finally, hyperparameter optimization.

\subsection{The Problem}
To begin with, our unknown function that describes the heat at time $t$ and at a spatial point
$x$ is $y(x, t)$ and the complete initial-boundary-value problem is:
\begin{align}
    \label{eq:heat}
    \begin{cases}
        \partial_t y(x, t) = \kappa \Delta y(x, t), \quad (x, t) \in [0, \pi] \times (0, 3] \\
        y(x, 0) = \sin(x) & \text{(Initial conditions)} \\
        y(0, t) = y(1, t) = 0 & \text{(Boundary conditions)}
    \end{cases}
\end{align}
where the Laplacian of $y$ is $\Delta y = \frac{\partial^2 y}{\partial x^2 } $,
and $\kappa$ is the thermal conductivity constant. Moreover, we know that Equation~\eqref{eq:heat} has
an exact solution in closed form: $y(x, t) = \sin(x) \exp(-\kappa t) $. Notice 
that Equation~\ref{eq:heat} has constant boundary conditions (Dirichlet), and the initial conditions
indicate that at time $t=0$, the heat spatial distribution is sinusoidal. 

\subsection{The Loss Function}

The loss function is the most important part of the Deep Galerkin method since it drives the learning of
the neural network. Therefore, we will show how to derive the loss function step-by-step:
First, we rewrite our PDE in the form: 
\begin{align}
    \label{eq:heat_loss_step1}
     \partial_t y(x, t) - \kappa \frac{\partial^2 y(x, t)}{\partial x^2} = 0.
\end{align}
Our neural network, $\mathcal{F}$ will approximate the solution $y(x, y)$, and thus it will have to 
get as closs as possible to Equation~\eqref{eq:heat_loss_step1}. Therefore, the first term of our loss
function is:
\begin{align}
    \label{eq:heat_loss_step1_nn}
     \partial_t \hat{y}(x, t) - \kappa \frac{\partial^2 \hat{y}(x, t)}{\partial x^2}.
\end{align}
Now, the question is: ``How do we impose the initial and boundary conditions?'' And the answer to that
is rather simple. We add extra terms in our loss that will penalize it in such a way to enforce the 
approximation, $\hat{y}$ to behave as it should at time $t=0$ (initial conditions, $\hat{y}(x, 0) = \sin(x)$)
and on the boundaries ($\hat{y}(0, t) = \hat{y}(1, t) = 0$). Therefore, the extra terms we have to add
to our loss function are: $(\hat{y}(x, 0) - y(x, 0))^2 = (\hat{y}(x, 0) - \sin(x))^2$ for penalizing 
the initial conditions, and $(\hat{y}(0, t) - y(0, t))^2 = \hat{y}(0, t)^2$ and $(\hat{y}(1, t) - y(1, t))^2 = \hat{y}(1, t)^2$ for the boundary conditions. 
Finally, since we use mini-batches of size $n$ to train our neural network, we will use the following notation
$\hat{{\bf y}}$ to indicate a one-dimensional vector $ \hat{{\bf y}}({{\bf x}, {\bf t})} = [\hat{y}(x_1, t_1), \ldots, \hat{y}(x_n, t_n)]^{\top}$,
where each pair $(x_i, t_i)$ is one sample out that belongs to a mini-batch. 
And now, we can write down our loss by averaging over all those mini-batches; thus, we have: 
\begin{align}
\label{eq:heat_loss}
    \mathcal{L}(\pmb{\theta}) &= \frac{1}{n} \sum_{i=1}^{n} \Big(\frac{\partial \hat{\Y}_i(\X_i, \Ti_i)}{\partial \Ti_i} - \kappa \frac{\partial^2 \hat{\Y}_i(\X_i, \Ti_i)}{\partial \X_i^2}\Big)^2 \nonumber \\
    &+ \frac{1}{n} \sum_{i=1}^{n} (\hat{\Y}_i(\X_i, 0) - \sin(\X_i))^2  \\
    &+ \frac{1}{n} \sum_{i=1}^{n} (\hat{\Y}_i(0, \Ti_i))^2 + \frac{1}{n} \sum_{i=1}^{n} (\hat{\Y}_i(1, \Ti_i))^2. \nonumber
\end{align}
Note that the last term in Equation~\eqref{eq:heat_loss} includes only the solution approximated by a
neural network on the boundaries. Since the boundary conditions are zero, we have nothing to subtract as we did for the initial conditions. 

\subsection{The Pytorch Loss Function}

Since we have our loss function, at least on paper, we can now proceed and implement a Pytorch 
function that will be used to train our neural network. The following code listing shows
the Pytorch implementation of the loss function based on Equation~\eqref{eq:heat_loss}.
The first step is to approximate the solution within the domain $\Omega$ using a neural network
(line $3$ of the snippet). Then, it computes the temporal and second spatial derivatives using Automatic
Differentiation. Finally, it computes all three terms of Equation~\eqref{eq:heat_loss} and returns the mean
over the batches. As you can see, the implementation is more or less a one-to-one mapping of our 
Equation~\eqref{eq:heat_loss} to Pytorch. 

\begin{lstlisting}[style=custompython,label=list:heat_loss,caption={{\bfseries \sffamily 1D Heat Equation Loss Function}}]
def heat1d_loss_func(net, x, x0, xbd1, xbd2, x_bd1, x_bd2):
    kappa = 1.0         # Thermal conductivity constant
    y = net(x)          # Obtain a neural  network approximation of heat eq. solution

    # Compute the gradient (derivatives) 
    dy = torch.autograd.grad(y,
                             x,
                             grad_outputs=torch.ones_like(y),
                             create_graph=True,
                             retain_graph=True)[0]
    dydt = dy[:, 1].unsqueeze(1)    # Get the temporal derivative
    dydx = dy[:, 0].unsqueeze(1)    # Get the spatial first derivative

    # Compute the second partial derivative
    dydxx = torch.autograd.grad(dydx,
                                x,
                                grad_outputs=torch.ones_like(u),
                                create_graph=True,
                                retain_graph=True)[0][:, 0].unsqueeze(1)

    # Compute the loss within the domain
    L_domain = ((dydt - kappa * dydxx)**2)

    # Compute the initial conditions loss term
    y0 = net(x0)
    L_init = ((y0 - torch.sin(x0[:, 0].unsqueeze(1)))**2)

    # Compute the boundary conditions loss terms
    y_bd1 = net(xbd1)
    y_bd2 = net(xbd2)
    L_boundary = ((y_bd1 - x_bd1)**2 + (y_bd2 - x_bd2)**2)

    # return the mean (over batches)
    return torch.mean(L_domain + L_init + L_boundary)
\end{lstlisting}

\subsubsection{The Neural Network}

The last piece we are missing is the neural network that will approximate our 
solution. We choose here to use a multilayer feed-forward neural network (MLP)
with two hidden layers of $32$ units each, initializing the parameters with
Xavier's uniform distribution, and we apply a $\tanh$ function as non-linearity
after each layer except the output. The architecture of the neural network is:

\begin{center}
    \FINL(2, 32) $\leadsto$ Tanh $\leadsto$ \FHL(32, 32) $\leadsto$ Tanh
    $\leadsto$ \FHL(32, 32) $\leadsto$ Tanh $\leadsto$ \FOL(32, 1).
\end{center}

As we can see, the neural network receives a two-dimensional input from the heat Equation~\eqref{eq:heat},
with temporal $t$ and spatial $x$ arguments. The neural network's output is one-dimensional and represents
the temperature obtained as a solution to the heat equation at point $x$ and time $t$. FC means fully
connected layer, and it implements an affine transformation as we already have seen
(${\bf y} = {\bf W} {\bf x} + {\bf b}$). In Pytorch, we define fully connected layers using the
\texttt{Linear} class. In addition, the output layers are fully connected, and only the input layer
serves as a placeholder.
Listing~\ref{list:mlp} provides the implementation of the MLP neural network used in this example.

\begin{practice}
    As you can see, we chose the MLP as an approximator in this example. You can try to replace 
    the MLP with a ResNet or a DGM (see Sections~\ref{section:resnet} and~\ref{section:dgm_net},
    respectively), and experiment with them to see how they behave. 
\end{practice}

\subsection{The Approximated Solution}

Before we proceed in seeing how the approximated solution looks like, and determine if our neural network
was able to approximate the solution we have to take care of a few last details. 
First, we have to determine the mini-batch size. In this case, we choose to set it to $64$, meaning 
that we sample from a uniform distribution both $t \sim \mathcal{U}(0, 3)$ and $x \sim \mathcal{U}(0, \pi)$
$64$ times and that mini-batch will be fed to the neural network for a predetermined number of iterations.
Keep in mind that at every iteration we obtain new samples for both $x$ and $t$ and we evaluate the 
loss function. In this example, we set the number of iterations to $5,000$.
The last piece is the optimizer, the variation of the gradient descent that will update our neural network's 
parameters. For this example, we use the Adam optimizer, which in Pytorch, can be imported from the
\texttt{optim}\footnote{\url{https://docs.pytorch.org/docs/2.12/optim.html}} package. In addition, the learning rate for this problem is set to $0.0001$ remains constant during training. 

Listing~\ref{list:heat_loss_loop} shows the training loop and thus how we can minimize the loss function in Pytorch. 
The function receives as inputs our neural network, the number of iterations, the batch size and the 
learning rate.

\begin{lstlisting}[style=custompython,label=list:heat_loss_loop,caption={{\bfseries \sffamily 1D Heat Equation Training Loop}}]
    def minimize_loss_dgm(net,
                      iterations=1000,
                      batch_size=32,
                      lrate=1e-4,
                      ):
    """! Main loss minimization function. This function implements the Deep
    Galerkin Method.

    @param net A torch neural network that will approximate the solution of
    neural fields.
    @param iterations Number of learning iterations (int).
    @param batch_size The size of the minibatch (int).
    @param lrate The learning rate (float) used in the optimization.

    @return A torch neural network (trained), and the training loss (list)
    """
    optimizer = torch.optim.Adam(net.parameters(), lr=lrate)

    t0 = torch.zeros([batch_size, 1], device=device)

    xbd1 = torch.zeros([batch_size, 1], device=device)
    xbd2x = torch.ones([batch_size, 1], device=device) * torch.pi
    xbd2y = torch.zeros([batch_size, 1], device=device)

    train_loss = []
    for i in range(iterations):
        x = torch.pi * torch.rand([batch_size, 1], device=device)
        t = 3.0 * torch.rand([batch_size, 1], device=device)

        X = torch.cat([x, t], dim=1)
        X.requires_grad = True

        X0 = torch.cat([x, t0], dim=1)

        X_BD1 = torch.cat([xbd1, t], dim=1)
        X_BD2 = torch.cat([xbd2x, t], dim=1)

        optimizer.zero_grad()

        loss = dgm_loss_func(net, X, X0, X_BD1, X_BD2, xbd1, xbd2y)

        loss.backward()
        optimizer.step()

        train_loss.append(loss.item())

        if i % 100 == 0:
            lr = optimizer.param_groups[0]['lr']
            print(f"Iteration: {i}, Loss: {loss.item()}, LR: {lr}")

    return net, train_loss
\end{lstlisting}

\begin{figure}[!htpb]
    \centering
    \includegraphics[width=1.\textwidth]{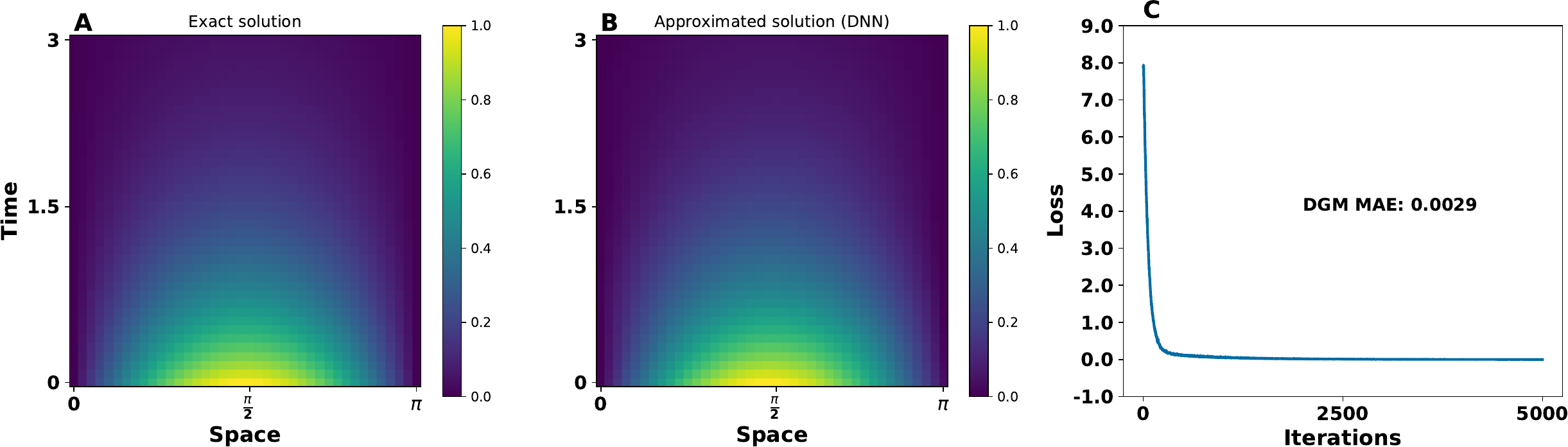}
    \caption{{\bfseries \sffamily $1$D Heat equation solution.}
    Panel {\bfseries \sffamily A} shows the exact solution of Equation~\eqref{eq:heat},
    {\bfseries \sffamily B} illustrates the approximated solution by the DG method and a
    deep feed-forward neural network, and panel {\bfseries \sffamily C} displays
    the loss over training iterations.
    }
    \label{fig:heat_sol}
\end{figure}

So we run the function provided in Listing~\ref{list:heat_loss_loop} and we 
we obtain the approximated solution shown in Figure~\ref{fig:heat_sol} panel B.
In the same figure, we see the exact solution (obtained analytically) in panel A.
By visually inspecting and comparing panels A and B, we observe that the two 
solutions are similar. When we compare thw two solution in a more rigorous and
systematic way by using the mean absolute error we confirm that the two solutions 
are close, $\text{MSE} = 0.003$.
Finally, in Figure~\ref{fig:heat_sol} C, we see the loss as a function of iterations,
and we note how it drops as we train our neural network on more and more samples,
meaning the neural network is learning to approximate the solution to our problem, 
and thus the loss is being minimized.

\subsection{Effect of Batch Size on Minimization}

Reading the previous paragraph, naturally, one may be wondering how we came up with 
the size of mini-batch or all the other hyperparameters (\emph{i.e.}, learning rate,
number of iterations, number of hidden layers and hidden units).
Let's see how we can determine the right size of 
a mini-batch for our problem, and later, we will see how to determine all the 
hyperparameters in an automated way. Moreover, while we are searching for the
proper mini-batch size we also see what effect that has on the minimization 
process. 

\begin{figure}[!htpb]
    \centering
    \includegraphics[width=0.6\textwidth]{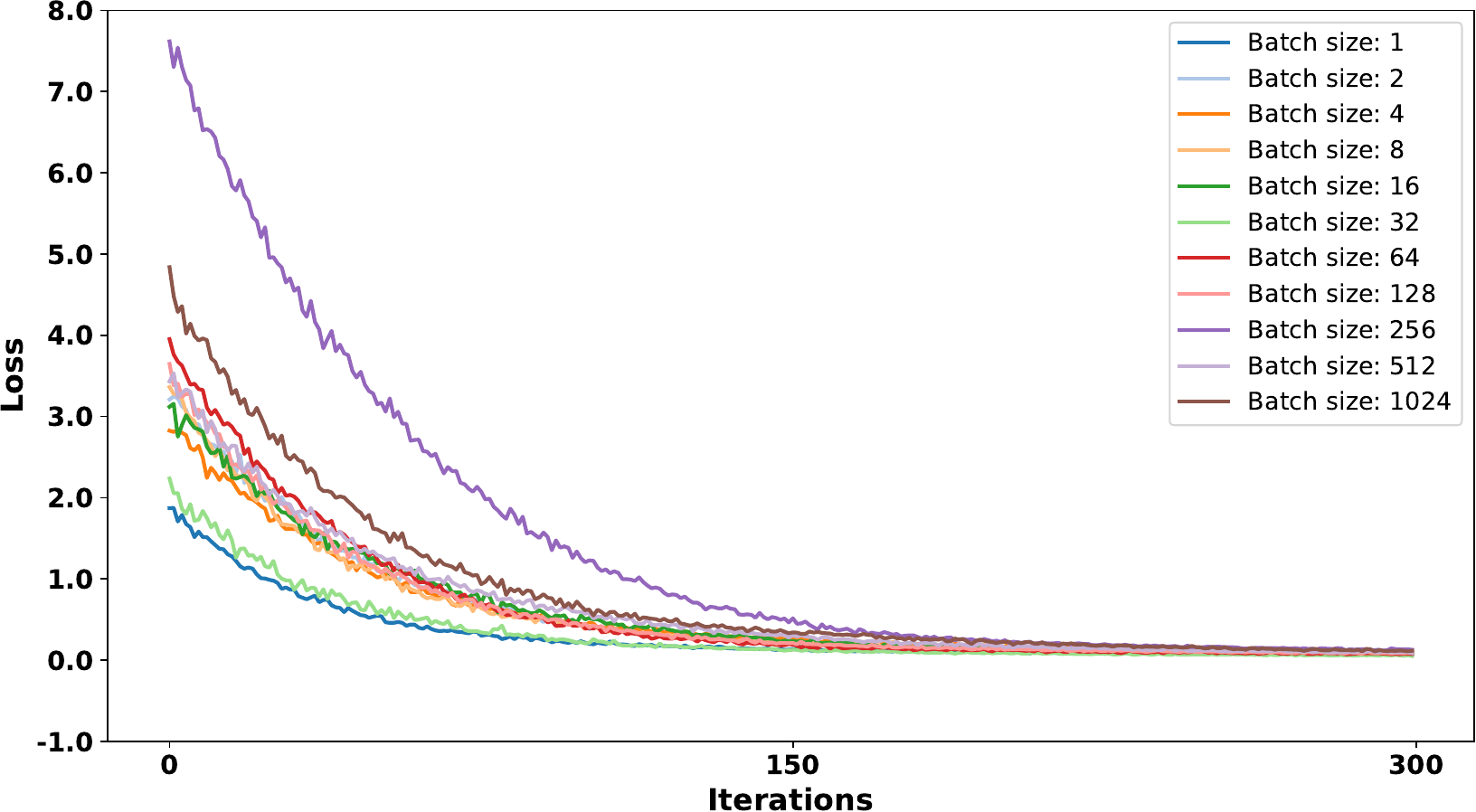}
    \caption{{\bfseries \sffamily Effect of Batch Size on Minimization} We see
    the evolution of the loss function~\eqref{eq:heat_loss} throughout $300$
    iterations when we solve the initial-boundary-value-problem described by
    Equation~\eqref{eq:heat}. Different batch sizes as powers of two have been
    used in this case ($n \in \{2^k | \, k \in \mathbb{N}\, \text{and} \, 0
    \leq k \leq 10 \}$). The inset shows the first $50$ iterations of the
    minimization problem. }
    \label{fig:batch_size}
\end{figure}

Let us consider the previously solved problem~\eqref{eq:heat}, keeping all hyperparameters
fixed except the mini-batch size. We systematically vary the mini-batch size from $2^0=1$ to $2^{10}=1024$,
solving the problem for $5,000$ iterations for each size, as before. To ensure accuracy, each experiment
is repeated $5$ times, and the mean is calculated across the 5 independent runs. Figure~\ref{fig:batch_size}
displays the mean loss over five experiments for the first $300$ iterations. The results clearly illustrate
the influence of mini-batch size on loss, allowing us to identify the batch size that optimally suits our problem.
In this context, our previously chosen batch size ($64$, deep red line) proves suboptimal for convergence 
speed; the loss decreases more slowly than with a mini-batch of $32$ (light green line) or even $16$ (dark 
green line) samples. 

\subsection{Effect of Batch Normalization on Minimization}

\begin{figure}[htpb!]
\centering
\includegraphics[width=.48\linewidth]{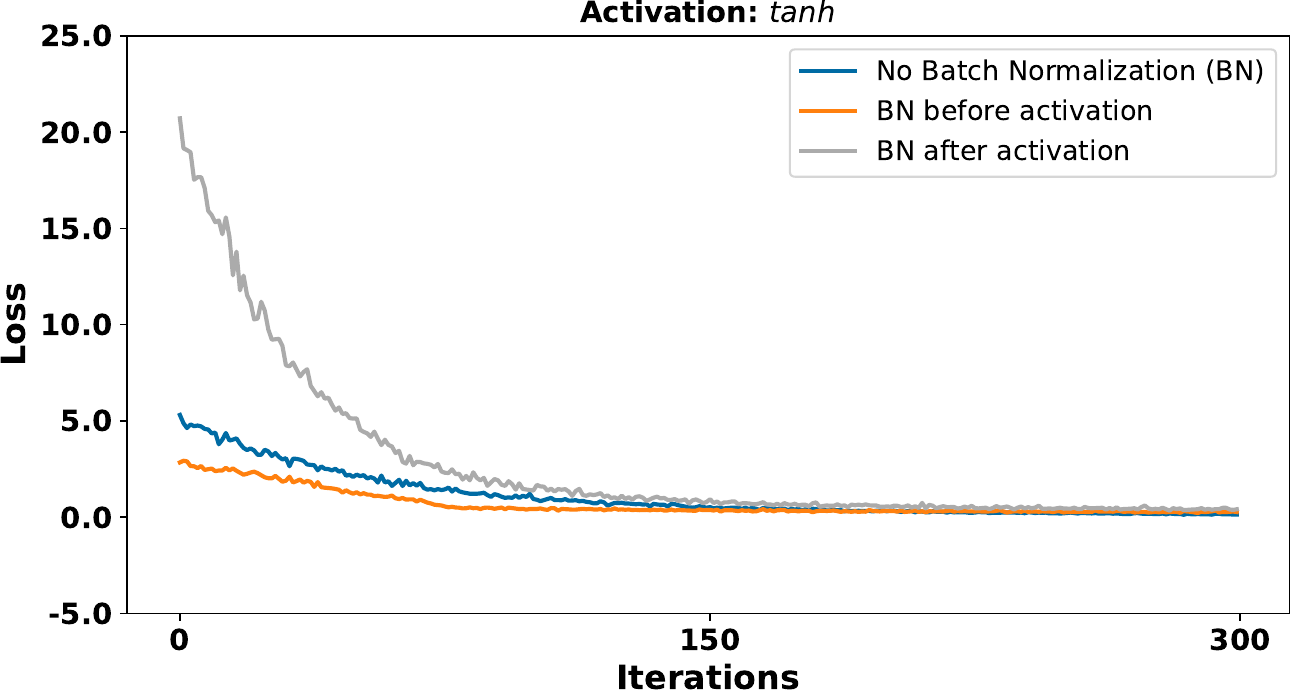}
\includegraphics[width=.48\linewidth]{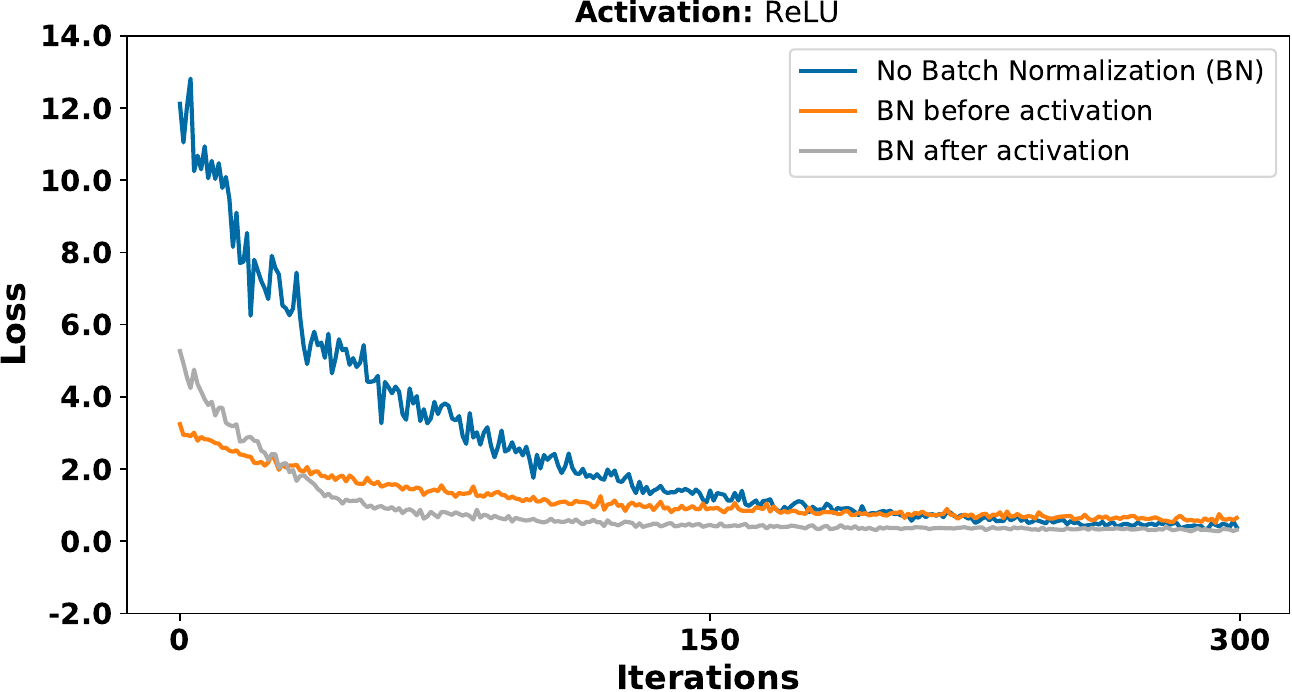}
\caption{{\bfseries \sffamily Effect of Batch Normalization on Minimization}
    The left panel shows the loss function (Equation~\eqref{eq:heat_loss}) over
    time for $300$ iterations when we solve the
    initial-boundary-value-problem described by Equation~\eqref{eq:heat}, and
    the activation function of the neural network is a $\tanh$ function. The
    blue line indicates the loss when we do not use batch normalization. The
    orange and gray lines show the loss when we use batch normalization before
    and after the non-linearity (activation function), respectively. The right
    panel shows the effect of batch normalization on the same neural network
    when we replace the $\tanh$ with a $ReLU$.}
    \label{fig:batch_norm}
\end{figure}

In the same spirit as before, we will try to figure out, in general, how to improve our learning process.
As we discussed, one way is to use batch normalization (see Section~\ref{section:bn}).
To this end, we apply batch normalization to our neural network layers, as shown in Listing~\ref{list:mlp}.
Then we try to run our little experiment as we did before. After training our neural network for $5,000$ iterations,
we can see in Figure~\ref{fig:batch_norm} (left panel) the average loss as a function of iterations averaged
over five independent experiments. Notice that when we apply batch normalization after the nonlinearity
($tanh$ — left panel or ReLU — right panel), the loss converges faster than in any other case
(no batch normalization or batch normalization before the activation function). 

\subsection{The Search for Hyperparameters ( ... and not for Spock)}

While we were trying to develop our neural network, we faced a few challenges regarding the hyperparameters (\emph{i.e.}, number of hidden layers and hidden units, number of iterations, mini-batch size, and learning rate) and how to decide their right values. Earlier, we did not explain how we arrived at the numbers we used; however, there is a \emph{smart} way to avoid that pitfall. We can use a Python library called Ray Tune to search the parameter space (\emph{i.e.}, where the hyperparameters live) with respect to the loss function, and by applying optimization algorithms and distributed computing techniques, it will find the appropriate set of hyperparameters for our problem. 

First, we should choose an optimization algorithm for Ray to use to search for the parameters. Here, we use the Optuna~\cite{akiba:2019} package to search over the number of iterations, batch size, and learning rate. Then we have to define a cost function, and that will be the heat loss function,
Equation~\eqref{eq:heat_loss}. 

Listing~\ref{list:objective} shows the Ray objective function for our problem,
which takes as an argument a Python dictionary \texttt{config} containing all the hyperparameters we are
optimizing (batch size, learning rate, and iterations). We, also, define a
neural network (lines $3$--$6$ in Listing~\ref{list:objective}), and we use
that neural network to approximate a solution of Equation~\eqref{eq:heat}.
Finally, we transmit the final loss value to Ray Tune (line $16$) for use in
the hyperparameter optimization process. In this case, we are trying to find
the set of hyperparameters that minimizes the actual loss. Nevertheless, we
could compute the MAE between the neural network's solution and the closed-form
solution. 

\begin{lstlisting}[style=custompython,label=list:objective,caption={{\bfseries \sffamily Objective function used with Ray Tune}}]
def objective(config):
   # Define the neural network
    net = MLP(input_dim=2,
              output_dim=1,
              hidden_size=128,
              num_layers=3).to(device)

    # Train the neural network to approximate the DE
    _, loss_dgm = minimize_loss_dgm(net,
                                    iterations=config["n_iters"],
                                    batch_size=config["batch_size"],
                                    lrate=config["lrate"],
                                    )

    # Report loss
    session.report({"loss": loss_dgm[-1]})
\end{lstlisting}

Finally, we have to plug the objective function into Ray's machinery to optimize it. Listing~\ref{list:ray} illustrates the procedure required to initiate the optimization process. 
First we have to define our search space (lines $3$--$6$), which consists of the batch size, number of iterations, and learning rate. At this point we define the distributions from which Ray will draw the samples
for the hyperparameters. 
Subsequently, the search algorithm is selected (line $9$); in this instance, the Optuna~\cite{akiba:2019} algorithm is chosen, along with the corresponding scheduler.
Finally, we instantiate the Tuner class (lines $14$--$26$) and pass it as
arguments: the resources we will use ($1$ GPU, if we are rich and have one at our disposal, and $10$ CPU cores).
We pass the objective function (\texttt{objective}) and the configuration, where we define
the metric (loss in this case) and a minimization. Moreover, we pass the Optuna
algorithm as an argument, the scheduler, and the number of samples for which
the algorithm will perform the hyperparameter space search. Finally, we pass
the search space (which is a Python dictionary), and we call the \texttt{fit}
method. One can find more details on using Ray Tune on its documentation
page\footnote{\url{https://docs.ray.io/en/latest/tune/index.html}}.

\begin{lstlisting}[style=custompython,label=list:ray,caption={{\bfseries \sffamily Hyperparameters tuning with Ray Tune}}]
def optimizeHeat():
    # Define the search space of hyperparameters
    search_space = {"batch_size": tune.randint(lower=1, upper=512),
                    "n_iters": tune.randint(1000, 50000),
                    "lrate": tune.loguniform(1e-4, 1e-1),
                    }

    # Set the Optuna optimization algorithm, and the scheduler
    algo = OptunaSearch()
    algo = ConcurrencyLimiter(algo, max_concurrent=5)
    scheduler = AsyncHyperBandScheduler()

    # Instantiate the Tuner class
    tuner = tune.Tuner(
                tune.with_resources(
                    objective,
                    resources={"cpu": 10,
                               "gpu": 1}),
                tune_config=tune.TuneConfig(metric="loss",
                                            mode="min",
                                            search_alg=algo,
                                            scheduler=scheduler,
                                            num_samples=10,
                                            ),
                param_space=search_space,
                )
    # Run the optimization
    results = tuner.fit()
\end{lstlisting}

\section{Ordinary Differential Equations}

In the previous section, we saw how to solve PDEs using the DG method. Importantly,
the DG method is equally effective in solving ordinary differential equations
(ODEs), systems of ODEs, and integral equations. Therefore, we continue this
primer by showing how to solve ODEs and systems of ODEs. 

\subsection{Exponential Decay} Our first example is about a first-order linear
ODE with initial conditions $y(0) = 2$, described by the following initial
conditions problem:
\begin{align}
    \label{eq:linear_ode}
    \begin{split}
           \frac{dy(t)}{dt} &= y(t), \\
            y(0) &= 2.
    \end{split}
\end{align}
Equation~\eqref{eq:linear_ode} admits an exponential solution, $y(t) =
2\exp(-t)$, that decays towards zero as time approaches infinity. Following the
same recipe we used for the heat equation, we first define a loss function to
solve Equation~\eqref{eq:linear_ode} with the DG method. We notice that
Equation~\eqref{eq:linear_ode} does not have any boundary conditions; thus, we
neglect the boundary condition term in equation \eqref{eq:dgm_loss}, and hence
we have
\begin{align}
    \label{eq:simple_loss}
       \mathcal{L}(\pmb{\theta}) &= \frac{1}{n} \sum_{i=1}^{n} \Big( \frac{d\hat{y}}{dt} + \hat{y} \Big)^2 
                                     + \underbrace{\frac{1}{n} \sum_{i=1}^{n} \Big( \hat{y}_0 - 2 \Big)^2}_{\text{initial conditions}},
\end{align}
where $\hat{y}$ is the output of the neural network $\mathcal{F}(\pmb{\theta};t)$.
In this case, the neural network receives a one-dimensional input, the
time $t$, and returns a one-dimensional output $y$. The neural network has one
hidden layers with $32$ neurons, one input layer, and one output layer
with one neuron each. The overall architecture is
\begin{center}
    \FINL(1, 32) $\leadsto$ Tanh $\leadsto$ \FHL(32, 32) $\leadsto$ Tanh $\leadsto$
    \FOL(32, 1).
\end{center}

\begin{lstlisting}[style=custompython,label=list:simple_ode_loss,caption={{\bfseries \sffamily Loss function for s simple ODE.}}]
def simple_ode_loss(y,  # Approximated solution (neural network)
                    y0, # Approximated solution (neural network) at t = 0
                    t,  # Input variable - time
                    y_ic):  # Initial condition value

    # Compute the temporal derivative
    dydt = torch.autograd.grad(y,
                               t,
                               grad_outputs=torch.ones_like(y),
                               create_graph=True,
                               retain_graph=True)[0]

    L_domain = ((dydt + y)**2)

    L_init = ((y0 - y_ic)**2)
    return torch.mean(L_domain + L_init)
\end{lstlisting}

We discretize the temporal dimension using $64$ time steps. The input to the
neural network is organized into mini-batches of size $n=64$ and thus each mini-batch
has a shape of $(n, 1)$. We populate the tensor at each iteration with $64$ data points
sampled from a uniform distribution $\mathcal{U}(0, 1)$ (since $t\in[0, 1]$).
By using mini-batches, we can accelerate the convergence of learning and speed
up the computations since we exploit GPU's architecture more efficiently due to
parallel processing of mini-batches and make it more
stable~\cite{goodfellow:2016}. We minimize the loss for $2,000$ iterations with
a fixed learning rate $\eta=0.0001$, and using the Adam optimizer. 
\begin{figure}[!htpb]
    \centering
    \includegraphics[width=0.8\textwidth]{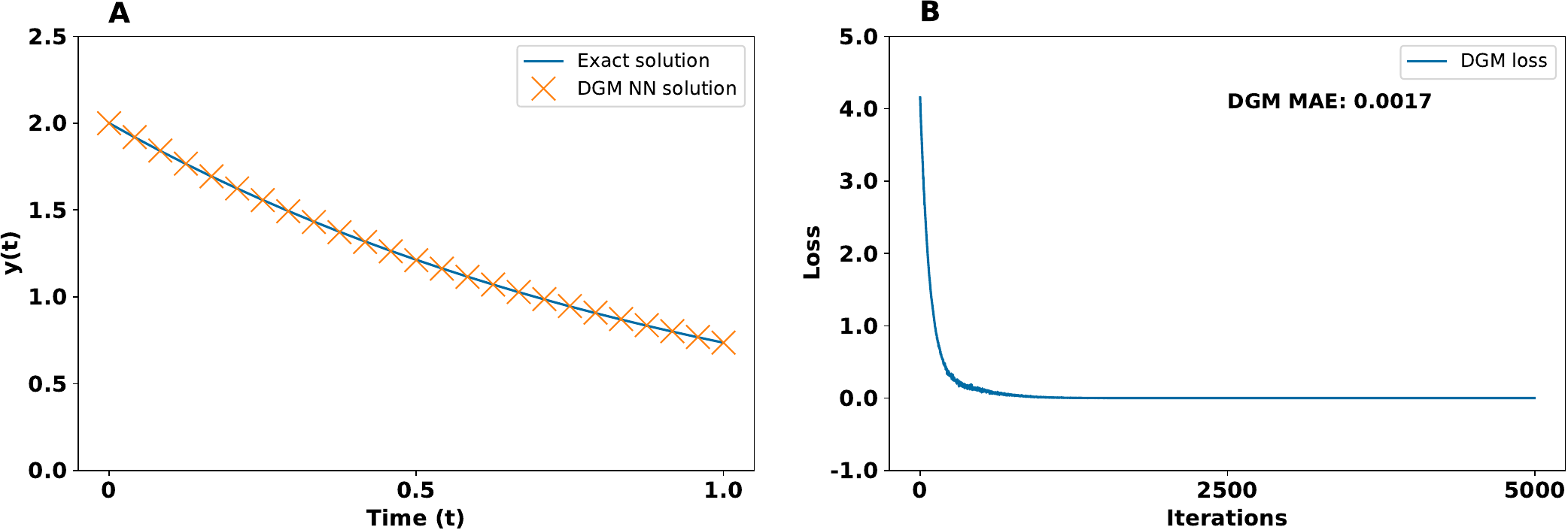}
    \caption{{\bfseries \sffamily First-order ordinary differential equation solution.}
    {\bfseries \sffamily A} Solution of Equation~\eqref{eq:linear_ode} in the interval $[0, 1]$.
    The blue solid line shows the analytic solution $y(t) = 2\exp(-t)$, and the orange crosses
    represent the solution obtained by the Deep Galerkin (DG) method. The approximated
    solution is close to the analytic one (mean absolute error MAE$=0.0017$).
    {\bfseries \sffamily B} Evolution of loss for the DG method. }
    \label{fig:first-ODE}
\end{figure}
Figure~\ref{fig:first-ODE} {A} shows the solution obtained by the neural
network in orange. We see that it is very close to the analytic solution (blue solid line),
something that we can confirm by estimating the MAE, which is $\text{MAE} = 0.0067$.
In addition, panel {C} shows the loss of the DG method. The DG
method converges to zero at about $1000$ iterations and stabilizes without 
oscillating or diverging.

\subsection{FitzHugh-Nagumo Model}

\begin{figure}[!htpb]
    \centering
    \includegraphics[width=\textwidth]{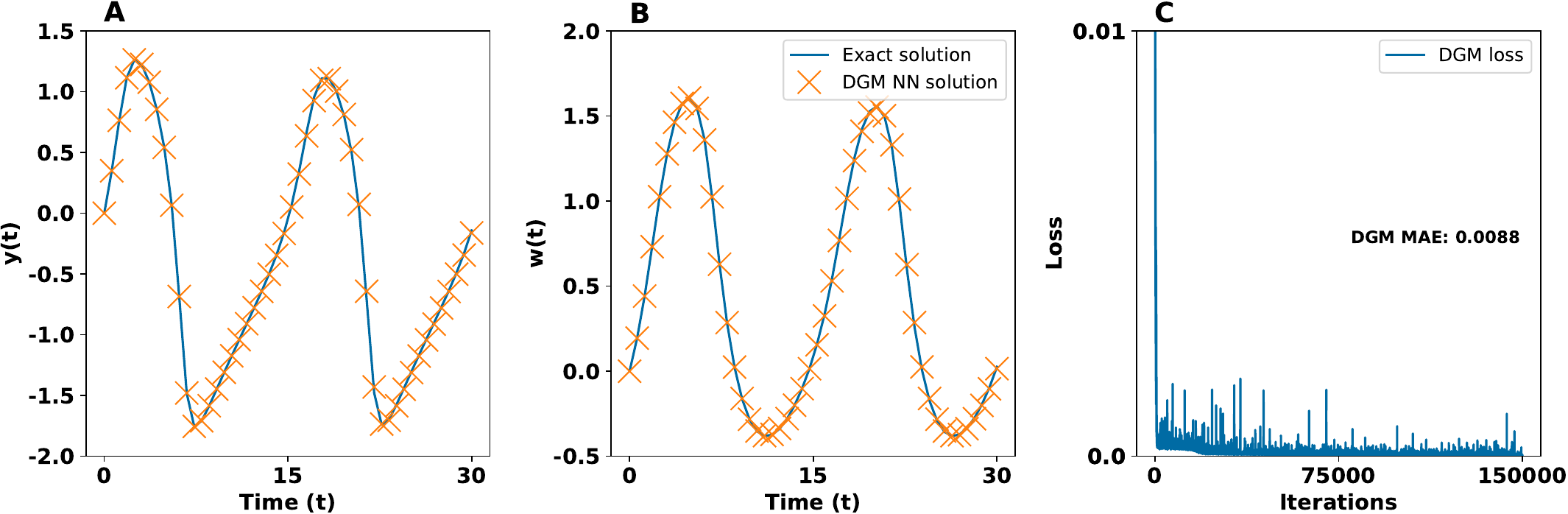}
    \caption{{\bfseries \sffamily FitzHugh-Nagumo model's solution.}
    {\bfseries \sffamily A} The solid blue line shows the solution $y(t)$ of
    equation~\eqref{eq:fitzhugh} solved numerically using \texttt{odeint} and the orange
    crosses indicate the approximated solution by a neural network of DGM-type
    (see main text). We see that the neural network approximates the solution
    provided by \texttt{odeint} well enough. The mean absolute error (MAE) is $\text{MAE}=0.0088$.
    {\bfseries \sffamily B} Similarly, the blue line indicates the numerical solution
    for $w(t)$ of Equation~\eqref{eq:fitzhugh}, and the orange crosses the solution
    obtained by the neural network.
    {\bfseries \sffamily C} Evolution of loss over learning iterations ($150,000$). The
    loss initially oscillates and, after about $90,000$, iterations converges with some
    small oscillations. Finally, after $150,000$ iterations, the loss has been minimized,
    and the neural network has learned a set of parameters $\pmb{\theta}$ that solves
    problem~\eqref{eq:fn_loss}.
    }
    \label{fig:nagumo}
\end{figure}

The second example is the FitzHugh-Nagumo
model~\cite{fitzhugh:1961,nagumo:1962}, a system of two coupled ordinary
differential equations (two-dimensional system) that describes the dynamics of
neuronal excitability. It reduces the Hodgkin-Huxley model of action potential
generation in the squid giant axon~\cite{hodgkin:1952}. The following equations
describe the FitzHugh-Nagumo model:
\begin{align}
    \label{eq:fitzhugh}
    \begin{split}
          \frac{dy(t)}{dt} &= y(t) - \frac{y(t)^3}{3} - w(t) + I_{\text{external}},  \\
           \tau \frac{dw(t)}{dt} &= y(t) + \alpha - \beta w(t), 
    \end{split}
\end{align}
where $y(t)$ is the membrane potential, $w(t)$ is a recovery variable, $I_{\text{external}}$
is the magnitude of the external current applied on the membrane, $\alpha$ and
$\beta$ are constants, and $\tau$ is a time decay constant. The variable $y$ allows
for regenerative self-excitation via positive feedback, and the $w$ follows linear
dynamics, as seen in equation~\eqref{eq:fitzhugh}, and provides a slower negative
feedback that acts as a recovery mechanism. Figure ~\ref{fig:nagumo} illustrates
a typical solution (blue line) of Equation~\eqref{eq:fitzhugh} when $\alpha = 0.7$,
$\beta = 0.8$, $\tau = 2.5$, and $I_{\text{external}}=0.5$. Panels {A}
and {B} show the $y$ and $w$ solutions in blue, respectively.
We solve numerically Equation~\eqref{eq:fitzhugh} using the \texttt{odeint} class
of \texttt{Scipy}. The initial conditions are $y(0) = 0$ and $w(0) = 0$, and we solve
for $t=30\, \mathrm{s}$ discretized with $50$ nodes.

On the other hand, we can solve the problem given by
Equation~\eqref{eq:fitzhugh} using the DG method. The major difference between
the current problem and the simple ODE of the previous example is that we have
to add an extra term in the loss function to describe the dynamics of $w(t)$
and adjust the output of the neural network since we have two dependent
variables, $y(t)$ and $w(t)$, in this case.  Therefore we proceed in defining
the loss function based on equation~\eqref{eq:dgm_loss}, and we get
\begin{align}
    \label{eq:fn_loss}
    \begin{split}
        \mathcal{L}(\pmb{\theta}) &= \frac{1}{n}\sum_{i=1}^{n} \Big( \frac{d\hat{y}}{dt} + ( \frac{\hat{y}^3}{3} + \hat{y} - I_{\text{external}} - \hat{w}) \Big)^2 \\
                                       &+ \frac{1}{n} \sum_{i=1}^{n} \Big( \frac{d\hat{w}}{dt} + (\beta \hat{w} - \alpha - \hat{y} ) / \tau \Big)^2 \\
                                       &+ \frac{1}{n} \sum_{i=1}^{n} \Big( \hat{{\bf s}}_0 - {\bf y}_{\text{IC}} ) \Big)^2, 
    \end{split}
\end{align}
where $\hat{y}$ and $\hat{w}$ are the solutions approximated by the neural
network for $y$ and $w$, respectively. The tensor ${\bf y}_{\text{IC}}$
contains the initial conditions with shape $(n, 2)$. The output tensor
of the neural network at time $t=0$ is $\hat{{\bf s}}_0$ of shape $(n,
2)$, where $n=256$ is the batch size. The neural network's output is
two-dimensional since we have two dependent variables, $y$ and $w$, and that's
why we use an extra tensor $\hat{{\bf s}}$ to store the output. In Pytorch we
assign $\texttt{y = s[:, 0]}$ and $\texttt{w = s[:, 1]}$. Generally, when we
have a $q$-dimensional system, the neural network's output would be a
$q$-dimensional tensor. Moreover, remember that the neural network takes a
one-dimensional input, the argument $t$ in equation~\eqref{eq:fitzhugh}. All
the parameters of the FitzHugh-Nagumo equations receive the same values as
before.

For this particular problem, we choose a DGM-type neural network with one input
unit that represents the time variable $t$, two output units that reflect the
solutions $y(t)$ and $w(t)$, and finally, four hidden DG layers with $128$
units each. We decided to go with a DMG-type neural network because of the
rapidly changing dynamics of Equation~\eqref{eq:fitzhugh}. Listing~\ref{list:dgm}
gives the implementation of the DGM layer used here.
Moreover, all the activation functions of the DGM neural network are linear
rectifiers (ReLU). The overall architecture of the neural network is
\begin{center}
    \FINL(1, 128) $\leadsto$ ReLU $\leadsto$ DGML(128, 128) $\leadsto$ DGML(128, 128)
    $\leadsto$ DGML(128, 128) $\leadsto$ DGML(128, 128) $\leadsto$  \FOL(128, 2).
\end{center}
Finally, we use the \texttt{grad} method of Pytorch to compute the temporal
derivatives $\frac{d\hat{y}}{dt}$ and $\frac{d\hat{w}}{dt}$. An alternative is
to use the \texttt{jacobian} method of the \texttt{autograd} since the neural
network's output is not scalar (one-dimensional). However, the
\texttt{jacobian} is generally slower than the \texttt{grad}, which is applied
on scalar neural network outputs. Thus, we separately apply the \texttt{grad}
method on each neural network output. Listing~\ref{list:fzng_loss}
below shows the implementation of the loss function for the FitzHugh-Nagumo
problem.

    \begin{lstlisting}[style=custompython,label=list:fzng_loss,caption={{\bfseries \sffamily Loss function for the FitzHugh-Nagumo system.}}]
def fitzhugh_nagumo_loss(y,     # Solution approximated by a neural network
                         y0,    # Solution approximated by a neural network at t = 0
                         t,     # Input variable - time
                         y_ic): # Initial conditions
    Iext = 0.5      # External current
    alpha, beta, tau = 0.7, 0.8, 2.5    # FitzHugh-Nagumo model parameters

    # Get the variables y, w
    Y, W = y[:, 0].unsqueeze(1), y[:, 1].unsqueeze(1)

    dY = torch.autograd.grad(Y,
                             t,
                             grad_outputs=torch.ones_like(Y),
                             create_graph=True,
                             retain_graph=True)[0]

    dW = torch.autograd.grad(W,
                             t,
                             grad_outputs=torch.ones_like(W),
                             create_graph=True,
                             retain_graph=True)[0]

    Lx = torch.sum((dY + (Y**3/3.0 - Y - Iext + W))**2)
    Ly = torch.sum((dW + (beta * W - alpha - Y) / tau)**2)
    L0 = torch.sum((y0 - y_ic)**2)

    return Lx + Ly + L0

\end{lstlisting}

The optimizer of choice here is the Adam optimizer with its default parameters.
We run the training loop for $150,000$ iterations, drawing at each iteration a
random sample from a uniform distribution $\mathcal{U}(0, 30)$ since we are
interested in solving for $t \in [0, 30]$. We start our training with a learning
rate $\eta = 0.0001$ and reduce it to $0.00001$ after $35,000$ iterations.
All the other parameters concerning Equation~\eqref{eq:fitzhugh} are the same
as the ones we previously used with the \texttt{odeint} solver. 

Figure~\ref{eq:fitzhugh} shows the solution we obtained after training the
neural network. We see in panels {A} and {B}
in orange color the solutions $y(t)$ and $w(t)$, respectively. We observe that the
numerical solution we obtained from the \texttt{odeint} solver matches the one we got
from the neural network, and the MAE is $0.0088$, meaning that the two solutions are
close. Panel {C} shows the loss function dropping to zero after
$10,000$ iterations and stabilizing after about $90,000$ iterations.  

\section{Fredholm Integral Equations}

Integral equations, as their name reveals, involve an unknown function $y:[a,b]
\rightarrow \mathbb{R}$, and its integrals. In their general form, integral
equations are described by the following equation:
\begin{align}
    \label{eq:integral}
    \alpha(x) y(x) &= g(x) + \lambda \int_{a(x)}^{b(x)} K(x, t) y(t) dt,
\end{align}
where $\alpha(x)$ and $g(x)$ are given functions (the function $g(x)$ usually
describes a source or external forces), and $\lambda$ is a parameter.  And the
function $K: [a, b] \times [a, b] \rightarrow \mathbb{R}$ is called the
integral kernel or just kernel. An integral equation described by
Equation~\eqref{eq:integral}, for which $\alpha(x) = 0$, and only the unknown
function $y$ appears under the integral sign is of the \emph{first kind},
otherwise it is the \emph{second kind}. In addition, if the known function
$g(x) = 0$, we call the integral equation \emph{homogeneous}. Finally, when the
integral limits $a(x)$ and $b(x)$ are constants, then the integral equation is
called the Fredholm equation; otherwise, it is called the Volterra equation.
For instance, the following integral equation is a Fredholm of the first kind: 
\begin{align}
    \label{eq:fredholm1}
    g(x) = \int_{a}^{b}K(x, t)y(t)dt.
\end{align}
And the following equation is an inhomogeneous Fredholm's equation of the second kind
\begin{align}
    \label{eq:fredholm2}
    y(x) &=  g(x) + \int_{a}^{b} K(x, t) y(t) dt.
\end{align}
We refer the reader to~\cite{polyanin:2008} for more information about integral
equations and their solutions.  For the rest of this section, we will focus on
Fredholm's equation of the second kind. We will solve
Equation~\eqref{eq:fredholm2} with a kernel function $K(x, t) =
\sin(x)\cos(t)$, and $g(x) = \sin(x)$. Thus we have
\begin{align}
    \label{eq:fredholm3}
    y(x) &=  \sin(x) + \int_{0}^{\frac{\pi}{2}} \sin(x) \cos(t) y(t) dt,
\end{align}
which admits an analytic solution $y(x) = 2 \sin(x)$.  Let's begin with
defining the loss function for the Equation~\eqref{eq:fredholm3}. Based on
Equation~\eqref{eq:dgm_loss} and Equation~\eqref{eq:fredholm3}, we get
\begin{align}
    \label{eq:fredholm_loss}
    \mathcal{L}(\pmb{\theta}) &= \frac{1}{n} \sum_{i=1}^{n} \Big( \hat{y} - \sin(x) - \mathcal{I}  \Big)^2,
 \end{align}
where $\hat{y}$ is the output of the neural network $\mathcal{F}(\pmb{\theta}, t)$
and $\mathcal{I}$ is the integral term in equation~\eqref{eq:fredholm3}. We will compute
the integral term using Monte Carlo integration similar to~\cite{guan:2022}. Hence, 
\begin{align}
    \label{eq:integral_term}
    \mathcal{I} &= \int_{a}^{b} K(x, t) y(t) dt \approx \frac{\#[a, b]}{k} \sum_{j=1}^{k} K(x_i, t_j) \hat{y}(t_j) = \frac{\#[0, \frac{\pi}{2}]}{50} \sum_{j=1}^{50} \sin(x_i) \cos(t_j) \hat{y}(t_j)
\end{align}
where $\#[a,b]$ is the length of $[a, b]$, which is $\frac{pi}{2}$ for
equation~\eqref{eq:fredholm3}, and the number of samples (random points) we
draw from a uniform distribution $\mathcal{U}(a, b) = \mathcal{U}(0,
\frac{\pi}{2})$ is $k=50$. So for every point $j \in k$, we draw a random
sample, we pass it to the neural network $\mathcal{F}$ to obtain the $\hat{y}$
and then we plug that in equation~\eqref{eq:integral_term} to get the integral
term $\mathcal{I}$. After we evaluate the integral term, we proceed in
minimizing the loss function~\eqref{eq:fredholm_loss}. Listing~\ref{list:fredholm_loss}
shows the Pytorch implementation of the loss
function for this problem, where in lines $8$-$12$, we evaluate the integral
part described by Equation~\eqref{eq:integral_term}, and in line $15$, we
compute the solution $\hat{y}$ using the DGM neural network. Finally, in
line $17$, we estimate the loss function based on
Equation~\eqref{eq:fredholm_loss}. 

\begin{boxK}
{\bf Monte Carlo Integration} It is a numerical method for estimating multidimensional integrals
such as $\mathcal{I} = \int_{\Omega} f({\bf x}) d{\bf x}$, where $\Omega \subseteq \mathbb{R}^d$.
The naive algorithm for the Monte Carlo integration consists of the following steps:
\begin{enumerate}
    \item Sample $k$ vectors ${\bf x}_1, {\bf x}_2, \ldots, {\bf x}_k$ from a uniform
    distribution on $\Omega$, ${\bf x}_j \sim \mathcal{U}(\Omega)$,
    \item Estimate the term $\frac{\#\Omega}{k} \sum_{j=1}^{k} f({\bf x}_j)$, where
    $\#\Omega = \int_{\Omega}d{\bf x}$.
\end{enumerate}
Due to the law of large numbers the sum eventually will converge to $\mathcal{I}$.
\end{boxK}

\begin{lstlisting}[style=custompython,label=list:fredholm_loss,caption={{\bfseries \sffamily Loss function for the second kind Fredholm equation.}}]
def fredhold_loss(net,  # Here we pass the neural network (Pytorch object) to the loss function
                  x,    # Independent variable 
                  k=50):    # Number of random points we use in the Monte Carlo integration

    # Monte Carlo integration
    dr = np.pi / (2 * k)        # Integration step
    
    integral = 0.0
    for i in range(k):
        t = np.pi/2.0 * torch.rand_like(x)
        integral += torch.sin(x) * torch.cos(t) * net(t)
    integral *= dr

    # Evaluate the neural network on input x 
    yhat = net(x)

    L = ((yhat - torch.sin(x) - integral)**2)
    return torch.mean(L)
\end{lstlisting}
The neural network, in this case, is a DGM (see Section~\ref{section:dgm_net}
and Listing~\ref{list:dgm}) with one unit input and one output unit since
equation~\eqref{eq:fredholm3} has one input
argument, $t$, and the function $y(t)$ returns a scalar output. One hidden
layer with $32$ units is sufficient to solve the problem. Again, we choose a
DGM since equation~\eqref{eq:fredholm3} involves trigonometric functions
that would cause sudden turns. ResNet is another type of neural network used
successfully in~\cite{guan:2022} to approximate solutions of integral equations
in high dimensions. In this work, we will show how a DGM can approximate
solutions of equation~\eqref{eq:fredholm3}. The architecture of the neural
network is
\begin{center}
    \FINL(1, 32) $\leadsto$ ReLU $\leadsto$ DGML(32, 32) $\leadsto$
    \FOL(32, 1).
\end{center}
We use the Adam optimizer with its default parameters and learning rate $\eta =0.0001$.
We train the neural network for $3,000$ iterations. In each iteration we uniformly draw
$n=32$ (batch size) random numbers from $[0, \frac{\pi}{2}]$ representing the independent
variable $x$ and we feed the mini-batch to the neural network.
\begin{figure}[!htpb]
    \centering
    \includegraphics[width=\textwidth]{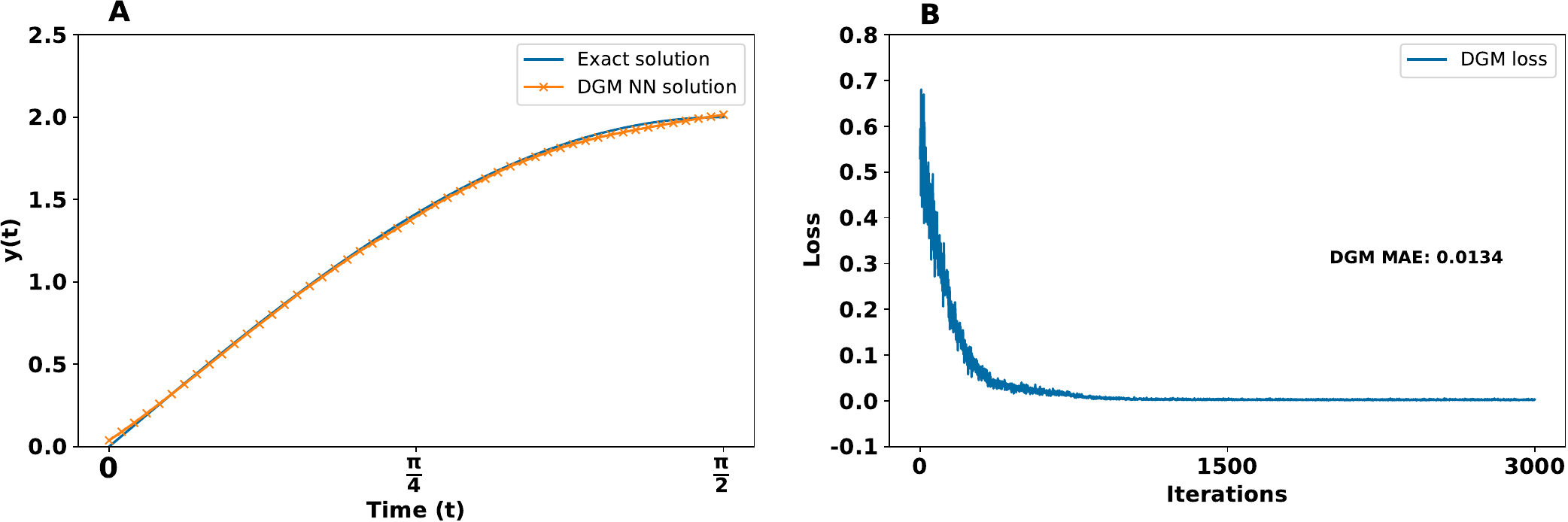}
    \caption{{\bfseries \sffamily Solution of a Fredholm equation of the second-kind.}
    {\bfseries \sffamily A} The blue curve is the analytic solution of equation~\eqref{eq:fredholm3},
    and the orange one is the approximated solution provided by a DGM-like neural network.
    The mean absolute error (MAE = $0.0134$) indicates that the two solutions are close.
    {\bfseries \sffamily B} The loss as a function of training iterations
    ($3,000$ total iterations) shows a transient phase for about $500$ iterations and then
    a convergence towards zero. 
    }
    \label{fig:fredholm}
\end{figure}
Figure~\ref{fig:fredholm} {A} shows the analytic solution of
Equation~\eqref{eq:fredholm3} (blue line) and the approximated solution evaluated with
the trained neural network on the interval $[0, \frac{\pi}{2}]$ with $50$ grid points.
We see that the approximated solution coincides with the analytic one, and the MAE=$0.0134$.
Panel~\ref{fig:fredholm} {B} shows the loss during training. We 
observe how the loss drops until convergence after about $500$ iterations.

\section{The End (plus a few more things we did not cover)}

This article provides an introduction to scientific deep learning by presenting one well-established method for solving partial differential equations. Here, we covered the basics of the Deep Galerkin and its practical aspects for implementing it in PyTorch. We saw step-by-step how to solve a one-dimensional
heat equation, and we provided all the necessary code snippets. Finally, we explained how we can search for the right hyperparameters based on the problem we want to solve. Moreover, we provided a methodology and practical tips that we can apply to other methods that use deep neural networks to solve differential equations (such as the DeepRitz method; see the next paragraphs). For a complete and more concrete review of many deep learning methods for differential equations, we refer the 
reader to~\cite{schrittwieser:2020,beck:2022}. In the next two paragraphs, we give a glimpse of what else is available in the field of scientific deep learning. Finally, we provide information about the source code accompanying this article and how to obtain it.

\subsection{Function-learning approaches}

This class of algorithms constitute the first major pillar of scientific deep learning, in which a neural network directly approximates the solution to a specific PDE.
Methods in this category — including Physics-Informed Neural Networks (PINNs)~\cite{raissi:2019}, the Deep Ritz Method~\cite{yu:2018}, and the Deep Galerkin Method~\cite{sirignano:2018} — use
neural networks to approximate PDE solutions by minimizing specially designed loss functions. PINNs are arguably the most widely adopted: they embed physical laws directly into the neural 
network architecture, enabling integration of domain knowledge so that models adhere to known physics while fitting available data, while the Deep Ritz method reformulates PDEs as energy minimization problems (drawing from the classical Ritz variational principle). Recent advances include residual-based adaptive sampling, improved activation functions, domain decomposition techniques, and hybrid auto-differentiation schemes, all aimed at enhancing accuracy and convergence in complex, high-dimensional, and multi-scale settings. A key limitation, however, is that function-learning approaches require training a separate neural network for each PDE or parameter configuration.

\subsection{Operator-learning approaches} This second class of algorithms addresses the limitation of function-learning approaches by learning a mapping between function spaces rather than
individual solutions, enabling generalization across entire families of problems. DeepONet~\cite{lu:2021} and Fourier Neural Operators~\cite{li:2021} achieved breakthrough performance by directly parameterizing solution operators, enabling zero-shot generalization across problem parameters and computational speedups of four to five orders of magnitude over classical numerical methods while maintaining comparable accuracy. Fourier Neural Operator parameterizes the integral kernel in Fourier space, while DeepONet decomposes the operator into a branch network (for input functions) and a trunk network (for query locations), both grounded in the universal approximation theorem. More recent work has hybridized these paradigms: physics-informed DeepONet and physics-informed FNO incorporate PDE-based loss functions into operator training, leveraging known physics within the operator-learning framework. 

\subsection{Source Code \& Various Implementations}
Last but not least, we provided many code snippets. Students can get actual working examples, including 
applications we did not cover here, such as solutions of ordinary differential equations (ODEs), systems of
ODEs, and solution of integral equations using deep neural networks. 
The endeavoring reader can find the source code on Github at \url{https://github.com/gdetor/differential_equations_dnn}. The code is licensed under the \texttt{GPL v3.0} license, and it is written in Python using PyTorch, NumPy, SciPy, Scikit-learn, Ray, and Matplotlib. On the same URL, you can find more details on the software/hardware specifications, as well as instructions on how to use the source code and running the examples. 

\bibliography{references}
\bibliographystyle{plain}

\end{document}